\pdfoutput=1
\documentclass[review,sort&compress,3p,12pt]{elsarticle}

\usepackage{mathtools,amssymb,xfrac,float,algpseudocode,algorithm}
\usepackage{graphicx,caption,subcaption,soul}
\usepackage{natbib,multirow}
\usepackage[bookmarks=false]{hyperref}
\usepackage[table]{xcolor}
\DeclareGraphicsExtensions {.pdf, .png , .jpg}
\newcommand{\RNum}[1]{\uppercase\expandafter{\romannumeral #1\relax}}
\usepackage{lineno}

\journal{Neurocomputing}

\begin{document}

\begin{frontmatter}

\title{Hyperspectral Image Classification Based on Sparse Modeling of Spectral Blocks}

\author[add1]{Saeideh Ghanbari Azar}
\ead{sghanbariazar@tabrizu.ac.ir}

\author[add1]{Saeed Meshgini\corref{cor1}}
\ead{meshgini@tabrizu.ac.ir}

\author[add1]{Tohid Yousefi Rezaii}
\ead{yousefi@tabrizu.ac.ir}

\author[add2]{Soosan Beheshti}
\ead{soosan@ee.ryerson.ca}
\address[add1]{Faculty of Electrical and Computer Engineering, University of Tabriz, Tabriz, Iran}
\address[add2]{Department of Electrical and Computer Engineering, Ryerson University, Toronto, ON, Canada}
\cortext[cor1]{Corresponding author}

\begin{abstract}
Hyperspectral images provide abundant spatial and spectral information that is very valuable for material detection in diverse areas of practical science. The high-dimensions of data lead to many processing challenges that can be addressed via existent spatial and spectral redundancies. In this paper, a sparse modeling framework is proposed for hyperspectral image classification. Spectral blocks are introduced to be used along with spatial groups to jointly exploit spectral and spatial redundancies. To reduce the computational complexity of sparse modeling, spectral blocks are used to break the high-dimensional optimization problems into small-size sub-problems that are faster to solve. Furthermore, the proposed sparse structure enables to extract the most discriminative spectral blocks and further reduce the computational burden. Experiments on three benchmark datasets, i.e., Pavia University, Indian Pines and Salinas images verify that the proposed method leads to a robust sparse modeling of hyperspectral images and improves the classification accuracy compared to several state-of-the-art methods. Moreover, the experiments demonstrate that the proposed method requires less processing time.
\end{abstract}


\end{frontmatter}

\section{Introduction}
\label{section.one}
Hyperspectral imagery collects energy scattered from a region in numerous spectral bands. Reducing the measurements to 3-10 spectral bands results in a coarser spectral resolution, which is called MultiSpectral Imagery (MSI). Multispectral images can be used to detect, for example soil areas in a region, whereas, HyperSpectral Images (HSIs) can, moreover, distinguish between minerals of the soil. The detailed data presented by a hyperspectral image offer significant benefits to material detection and classification in numerous fields, including agriculture \cite{gowen07}, astronomy \cite{courbot17}, biomedical imaging \cite{lu14,wei2019}, etc.

Although advantageous, this wealth of spectral data can be a serious challenge for acquiring, transmitting, and processing of HSIs. Moreover, these challenges often restrict the spatial resolution of HSIs, which lead to spectral mixing phenomenon in hyperspectral pixels. The Linear Mixture Model (LMM) is often used to describe this phenomenon and it will be mentioned in next section.

HSI classification is an important task where it is assumed that each pixel belongs to a specific class and the spectral mixing is not considered. Classification attempts to assign a specific label to each pixel. Early studies on HSI classification focus on spectral information and classify the spectral signature of the pixels using typical classifiers like SVM or neural networks \cite{ratle2010semisupervised}. Due to the spectral redundancy of the data, dimension reduction or band selection techniques can also be utilized to deal with high spectral dimensions \cite{bai2015semisupervised,guo2006band}. These methods are not very successful in reaching high accuracies because they consider each pixel separately and ignore the valuable spatial correlation of the pixels. Various image processing based methods such as Local Binary Pattern (LBP)\cite{li2015local}, wavelet transforms \cite{hsu2007feature} and morphological profiles \cite{dalla2010classification} have been successfully used in many recent studies to extract spatial features.

Deep learning based methods have also achieved remarkable results for hyperspectral classification with the cost of high computational time \cite{chen2016deep,xu2018spectral,zhong2017spectral,xu2018hyperspectral}. Convolutional Neural Networks (CNNs) are among the most popular networks for HSI classification. Chen et al use 3D CNNs with different regularization to extract joint spectral-spatial features \cite{chen2016deep}. Some other studies use a cascade of networks to extract separate spectral and spatial features. For instance \cite{zhong2017spectral} use two consecutive residual networks to extract spectral and spatial features and \cite{xu2018spectral} use LSTM and CNN models for spectral and spatial feature extraction, respectively. Two main challenges that deep learning based methods deal with include high computational time requirement and limited number of training data.

Sparse Representation (SR) is among the many approaches that have successfully been utilized for hyperspectral image processing. It solves the classification problem of HSIs using the LMM \cite{iordache11}. The extensive use of the SR in HSI processing stems from its excellent theoretical base \cite{donoho06} and successful performance in diverse areas of signal processing \cite{bofill01} and machine vision applications \cite{wright09,keinert2019}. It aims at finding a compact, yet efficient representation of a signal by expressing it as a linear combination of few atoms selected from a dictionary.

Some SR-based approaches make use of a priori available dictionary. For instance, standard spectral libraries or a dictionary constructed from training data have been used in many studies \cite{iordache11,chen11}. As an example, Peng et al. \cite{peng2019} use a dictionary, which is constructed from training samples and their spatial neighbors. To achieve a better representation of each test set, they build a local adaptive dictionary by selecting the most correlated atoms of the original dictionary to the test set. Although using a fixed dictionary has the advantage of small computational burden, it suffers from certain drawbacks. For example, using training data as dictionary makes the developed method strongly dependent on the choice of training data or using spectral libraries require an extra calibration stage \cite{bioucas12}. Recent studies take a step forward and train an optimal dictionary \cite{soltani15,fu18,xie2018}. SR with a trained dictionary or sparse modeling increases the performance of the HSI processing approaches with the cost of the increased computational burden \cite{castrodad11}.

One important factor affecting the performance of HSI processing is the strategy that a method uses to exploit the spectral and spatial information. Several previous studies have concentrated on spectral information and ignored the spatial correlations of the neighboring pixels \cite{camps05,melgani04}. Recent studies, on the other hand, take advantage of this spatial correlation and improve the performance. They usually define spatial groups and impose a certain constraint on the sparse coefficients of the defined spatial groups. For instance, Soltani et al. \cite{soltani15} define square non-overlapping patches as spatial groups. Assuming that the pixels inside a spatial group often belong to the same class, they constraint the sparse coefficients of these pixels to have a common sparsity pattern. Fu et al. \cite{fu18} use the same spatial groups and impose Laplacian constraint \cite{gao13} on the sparse coefficients of the spatial groups. Huang et al. \cite{Huang2019} use sliding square patches for unmixing of the HSIs. These fixed-size spatial groups are easy to implement and efficient in exploiting the spatial correlations of the pixels. Nevertheless, some other studies define more adaptive spatial groups \cite{fu2015,he16}. For instance, Zhi He et al. \cite{he16} utilize super-pixel segmentation to obtain homogenous regions and use them as spatial groups.

Various other methods have been developed for HSI processing with their specific pros and cons. However, there are still some challenges remained, two of which this study aims to address. First, although the dimensions of the data are high for HSIs, there are spectral and spatial redundancies that can be exploited. Various studies have proposed different strategies, but fully and jointly exploiting these redundancies remains a challenge. Second, since the complexity of the SR-based optimization problems depend strongly on the dimensions of the data, the high dimensions of the HSIs lead to computationally expensive optimization problems. This study introduces spectral blocks and proposes to use spectral blocks along with spatial groups to jointly exploit spectral and spatial correlations. Using these spectral blocks, this study divides the high-dimensional optimization problems into small-size sub-problems that are easier to tackle. Furthermore, the proposed structure provides the opportunity to further exploit spectral redundancy, determine the most discriminative spectral blocks, and discard the rest. This further reduces the computational burden and leads to more stable sparse coefficients and therefore, better classification accuracies with less processing time.

The remainder of this paper contains the following sections: \hyperref[section.two]{Section \ref*{section.two}} presents a brief overview of the sparse representation based HSI classification. \hyperref[section.three]{Section \ref*{section.three}} describes the proposed method. \hyperref[section.four]{Section \ref*{section.four}} presents the experimental results and the paper is concluded in \hyperref[section.five]{Section \ref*{section.five}}.
\section{Background and Preliminaries}
\label{section.two}
\begin{table}
  \centering
  \caption{List of the notations used in this paper.}\label{table.notations}
  \resizebox{0.85\textwidth}{0.25\textheight}{%
\begin{tabular}{|c|c|}
  \hline
  Notation & Description \\
  \hline
  $S$ & Number of spectral bands\\ \hline
  $N$ & Number of pixels of the image\\ \hline
  $G$ & Number of spatial groups in the image\\ \hline
  $m\times m$ & Size of the spatial groups or spatial windows\\ \hline
  $m^2$ & Number of pixels in a spatial group\\ \hline
  $B$ & Number of spectral blocks\\ \hline
  $s$ & Number of spectral bands in a spectral block\\ \hline
  $\mathbf{Y}_i \in \mathbb{R}^{S \times m^{2}}$ & i'th spatial group\\ \hline
  $\mathbf{Y}_{ij} \in \mathbb{R}^{s \times m^{2}}$ & j'th spectral block of the i'th spatial group\\ \hline
  $\mathbf{Y}_s \in \mathbb{R}^{s \times BN}$ & Observation matrix constructed for training the sub-dictionary\\ \hline
  $\mathbf{X}_i \in \mathbb{R}^{Bk \times m^{2}}$ & Sparse coefficient matrix corresponding to $\mathbf{Y}_i$ \\ \hline
  $\mathbf{X}_{ij} \in \mathbb{R}^{k \times m^{2}}$ & Sparse coefficient matrix corresponding to $\mathbf{Y}_{ij}$ \\ \hline
  $\mathbf{D}\in \mathbb{R}^{s \times k}$ & Sub-dictionary \\ \hline
  $\mathbf{A}\in \mathbb{R}^{S \times K}$ & Dictionary ($\mathbf{A}=\mathbf{I}\otimes \mathbf{D}$) \\ \hline
  $k$ & Number of atoms of the sub-dictionary $\mathbf{D}$\\ \hline
  $K=Bk$ & Number of atoms of the dictionary $\mathbf{A}$\\ \hline
  $\mathbf{I}\in \mathbb{R}^{B \times B}$ & Identity matrix of size $B$ \\ \hline
  $\mathbf{W}\in \mathbb{R}^{B \times B}$ & A binary diagonal matrix replaced with $\mathbf{I}$ to have selective spectral bands \\ \hline
\end{tabular}
}
\end{table}
In this section, we shortly review the spectral mixing of hyperspectral data formulated as a sparse modeling problem. In the following, boldface uppercase letters and boldface lowercase letters are used for matrices and vectors, respectively, while regular letters denote scalars. \hyperref[table.notations]{Table \ref*{table.notations}} presents a list of the notations used in this paper.

The LMM considers the spectrum of each pixel as a weighted linear combination of pure material spectra, referred to as \emph{endmembers}, with the weights being the corresponding \emph{abundances}. Assuming $N$ total number of pixels with $S$ spectral bands and stacking their spectra in a matrix, the LMM can be written as:
\begin{equation}\label{Eq.2}
  \mathbf{Y}=\mathbf{A}\mathbf{X}+\mathbf{E}
\end{equation}
where $\mathbf{Y}=\left[\mathbf{y}_{1},\mathbf{y}_{2},…,\mathbf{y}_{S}\right]\in \mathbb{R}^{(S\times N)}$ is the stacked pixel spectra in a matrix format and $\mathbf{A}\in \mathbb{R}^{(S\times K)}$ is a dictionary of $K$ atoms. $\mathbf{X}\in \mathbb{R}^{(K\times N)}$ and $\mathbf{E}\in \mathbb{R}^{(S\times N)}$ denote the abundance and error matrices, respectively. The elements of the error matrix are assumed to be i.i.d. random variables from a Gaussian distribution. This matrix form allows collaboratively manipulating the pixels and taking into account their spatial correlations.

Assuming that $\mathbf{A}$ contains a large number of available endmembers, for each pixel a small number of these endmembers will be mixed in its spectrum resulting in a sparse abundance vector. This is where sparse representation appears to find an optimal subset of the library atoms that can best represent the mixed spectrum of the pixel. Sparse representation aims at reducing the representation error of each pixel while inducing sparsity in the coefficient vector. Thus, assuming a Gaussian noise, the sparse representation problem can be written as the constrained form of:
\begin{equation}\label{Eq.3}
 \hat{\mathbf{X}}=\underset{\mathbf{X}}{\operatorname{argmin}}\, \, \left\{ \frac{1}{2}{\lVert \mathbf{Y}-\mathbf{A}\mathbf{X} \rVert}_{F}^{2}+\mu \,\textrm{R}(\mathbf{X})\right\}
\end{equation}
where $\textrm{R}(\mathbf{X})$ is a regularizer which mainly induces a form of desired sparsity pattern on $\mathbf{X}$ and $\mu$ is the regularization parameter controlling the importance of the regularization term. When the sparse coefficient vector of each pixel is calculated, a conventional classifier such as SVM can then be trained to classify the HSI pixels. This is the basic paradigm of sparse representation-based HSI classification.

The success of the above-mentioned method relies strongly on the choice of the dictionary matrix. Recent studies tend to learn an optimal dictionary adapted to the data rather than using a pre-defined fixed dictionary. In order to learn the dictionary, the following optimization problem should be solved:
\begin{equation}\label{Eq.4}
  \left\{ \hat{\mathbf{X}}, \hat{\mathbf{A}} \right\}= \underset{\mathbf{X},\mathbf{A}}{\operatorname{argmin}}\, \, \left\{ \frac{1}{2}{\lVert \mathbf{Y}-\mathbf{A}\mathbf{X} \rVert}_{F}^{2}+\mu \,\textrm{R}(\mathbf{X})\right\}\, .
\end{equation}
Assuming a convex regularizer, this problem becomes convex in either $\mathbf{X}$ or $\mathbf{A}$ but not in both. The general approach to tackle this optimization problem is to break it into two convex sub-problems and solve it in an iterative manner. The main two steps include \emph{sparse inference} which fixes $\mathbf{A}$ and optimizes $\mathbf{X}$ and \emph{dictionary update} which updates the dictionary with a fixed $\mathbf{X}$.

\section{The Proposed Method}
\label{section.three}
The amount of computations for the aforementioned optimization problems depends on the dimension of the data. Therefore, the high dimensions of the hyperspectral data lead to computationally expensive optimization problems. The existing methods for hyperspectral image classification solve these high-dimensional problems. This study attempts to reduce the required computations by jointly exploiting the spatial and spectral redundancy of the data. In addition to the commonly used spatial groups, we propose to exploit the spectral redundancy of the data and define spectral blocks. We simultaneously use the spatial groups and spectral blocks to divide the aforementioned optimization problems into small-size sub-problems. That is, instead of solving the global high-dimensional problem, we break it into several low-dimensional sub-problems which are computationally less expensive to solve. This proposed sparse structure also offers the advantage of requiring a low-dimensional dictionary, which notably reduces the computations of the dictionary-learning step. Moreover, this structure provides the opportunity of viewing the hyperspectral image in a multispectral resolution and discard the less effective spectral blocks resulting in increased stability of sparse coefficients and further reduction of the computational burden.
\subsection{Spectral Blocks}
\begin{figure}
 \centering
 \includegraphics[width=0.9\textwidth]{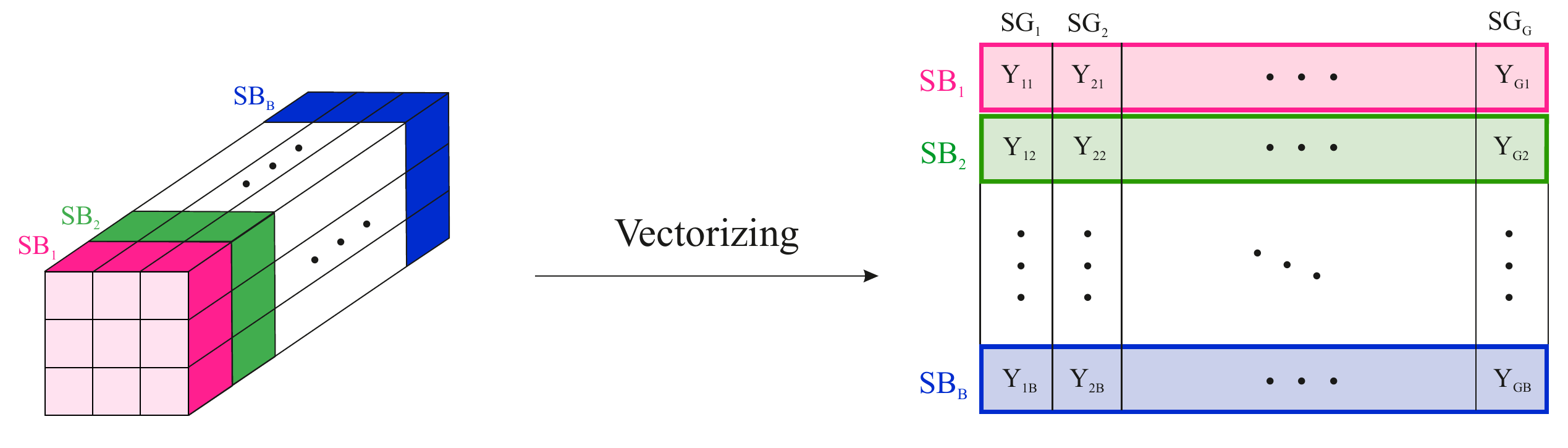}
 \caption{The division of the hyperspectral image into Spectral Blocks (SBs) and Spatial Groups (SGs). The image is then vectorized to be prepared for sparse modeling. Different colors represent different spectral blocks.}\label{fig.hsi}
\end{figure}
Different spatial groups have been utilized in many HSI classification methods to increase the stability of the sparse model \cite{fu18}. In this study, the spatial dimension of the data is divided into $G$ square ($m\times m$) spatial groups as in \cite{soltani15}. In addition, we propose to exploit the correlation among the spectral bands and divide the spectral dimension of the data into spectral blocks. For the sake of simplicity and speed, the spectrum of the data is divided into $B$ equal blocks, each block containing $s$ spectral bands. Nevertheless, for future research one may use different methods to obtain smarter spectral blocks. \hyperref[fig.hsi]{Figure \ref*{fig.hsi}} illustrates how the HSI is divided into spatial groups and spectral blocks. It also presents the vectorized form of the data. In this figure, different colors represent different spectral blocks. $\{\mathbf{Y}_{i}\}_{i=1,…,G}$ denotes the set of spatial groups and for the $i$’th spatial group, the spectral blocks are defined as $\{\mathbf{Y}_{ij}\}_{j=1,…,B}$.
\subsection{Sparse Modeling of Spectral Blocks(SMSB)}
For the $i$'th spatial group, the LMM can be written as:
\begin{equation}\label{Eq.5}
  \mathbf{Y}_{i}=\mathbf{A}\mathbf{X}_{i}+\mathbf{E}_{i}.
\end{equation}
Using the spectral blocks, we define the dictionary A as:
\begin{equation}\label{Eq.6}
 \mathbf{A}=\mathbf{I}\otimes\mathbf{D}
\end{equation}
where $\mathbf{I}_{B\times B}$ is the identity matrix, $\mathbf{D} \in \mathbb{R}^{(s\times k)} $ is a sub-dictionary of $k$ atoms and $\otimes$ denotes the matrix Kronecker product. The structure of $\mathbf{A}$ can be written as:
\begin{equation}\label{Eq.7}
  \mathbf{A}=\left[
    \begin{array}{ccc}
      \mathbf{D}& \cdots & 0 \\
      \vdots & \ddots & \vdots \\
      0 & \cdots & \mathbf{D} \\
    \end{array}
  \right].
\end{equation}

The first step is to learn the sub-dictionary D. For this end, the spectral blocks of all spatial groups are stacked together to form the following observation matrix:
\begin{equation}\label{Eq.8}
  \mathbf{Y}_{s}=\left[\mathbf{Y}_{11}, ...,\mathbf{Y}_{1B}, \mathbf{Y}_{21}, ..., \mathbf{Y}_{2B}, ..., \mathbf{Y}_{G1},...,\mathbf{Y}_{GB}\right],
\end{equation}
where $\mathbf{Y}_{s}$ has $s$ rows and $BN$ columns. Refer to \hyperref[fig.hsi]{Figure \ref*{fig.hsi}} and consider the Indian pines image with 200 spectral bands ($S=200$) as an example. If we divide the spectral dimension of this image into 10 blocks ($B=10$ and $s=20$) and stack the spectral blocks of all the spatial groups, the resulting $\mathbf{Y}_s$ would have $s=20$ rows and $10N$ columns. Using this constructed observation matrix, $D$ is trained by the following optimization problem:
\begin{equation}\label{Eq.9}
   \hat{\mathbf{D}}= \underset{\mathbf{X}_{s},\mathbf{D}}{\operatorname{argmin}}\, \, \left\{ \frac{1}{2}{\lVert \mathbf{Y}_{s}-\mathbf{D}\mathbf{X}_{s} \rVert}_{F}^{2}+\mu \,{\lVert \mathbf{X}_{s} \rVert}_{1}\right\}\, .
\end{equation}
This problem is solved using the online dictionary learning algorithm of \cite{mairal10} implemented by the SPAMS toolbox \cite{mairal14}.

\begin{figure}
 \centering
 \includegraphics[width=0.7\textwidth]{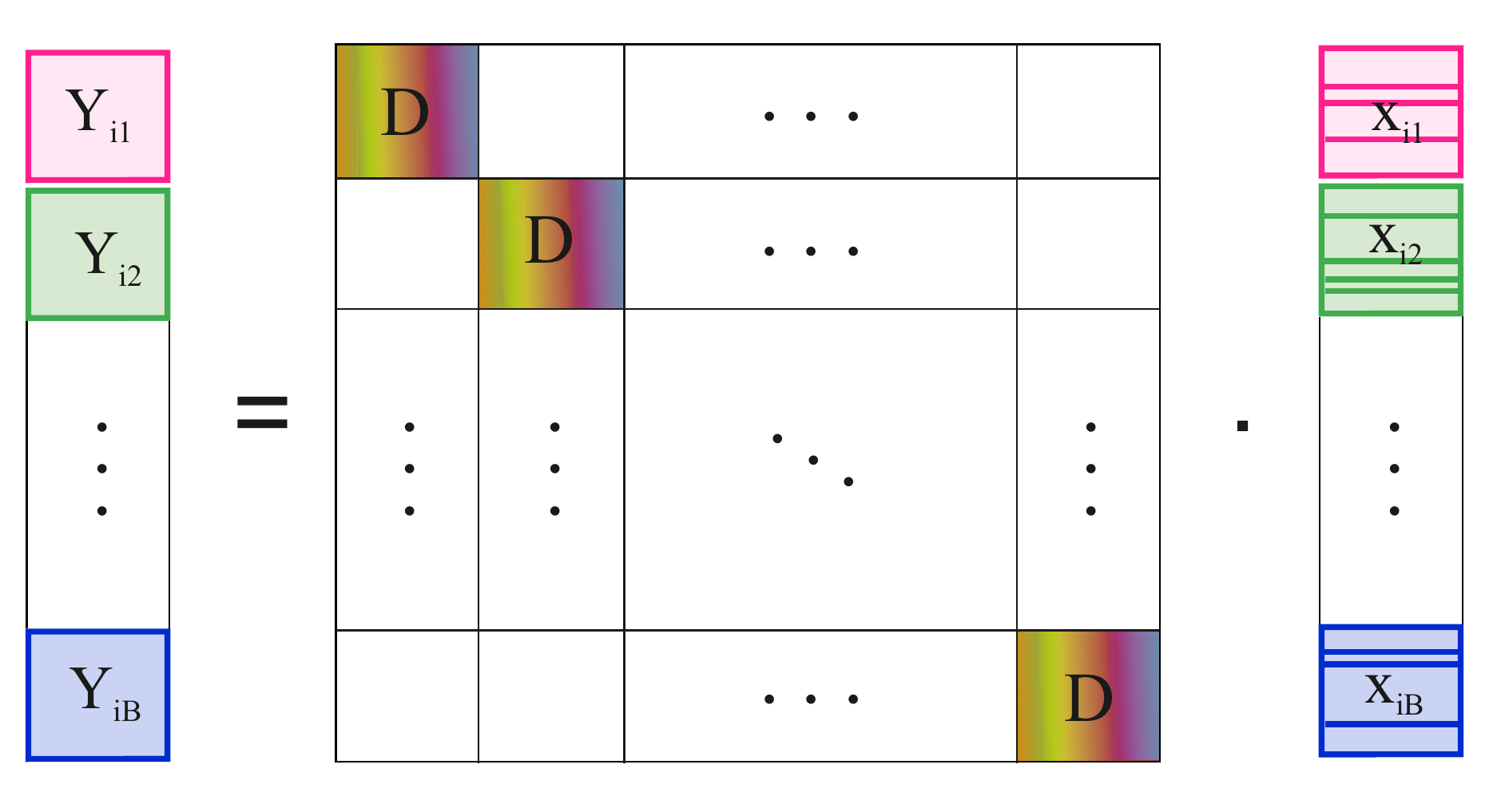}
 \caption{The sparsity pattern of the proposed SMSB method.}\label{fig.pattern}
\end{figure}

In the second step, using the trained sub-dictionary the sparse representation problem takes the following form for the $i$'th spatial group $\mathbf{Y}_i \in \mathbb{R}^{(S\times m^2)} $:
\begin{equation}\label{Eq.10}
 \underset{\mathbf{X}_{i}}{\operatorname{argmin}}\, \, \left\{ \frac{1}{2}{\lVert \mathbf{Y}_{i}-(\mathbf{I}\otimes \mathbf{D})\mathbf{X}_{i} \rVert}_{F}^{2}+\mu \,{\lVert \mathbf{X}_{i} \rVert}_{2,1}\right\}
\end{equation}
Here, the $\ell 2,1$-norm is used as a regularizer to imposes row-sparsity on the coefficient matrix of $\mathbf{X}_{i}$. \hyperref[fig.pattern]{Figure \ref*{fig.pattern}} illustrates the sparsity pattern of the proposed structure.

The advantage of the proposed sparse structure is twofold. First, instead of learning a large global dictionary, this method learns a small-size sub-dictionary which notably reduces the computational burden of dictionary-learning stage. Second, this block structure enables us to break the high-dimensional problem of \eqref{Eq.10} into $B$ small-size problems of the form:
\begin{equation}\label{Eq.11}
 \underset{\mathbf{X}_{ij}}{\operatorname{argmin}}\, \, \left\{ \frac{1}{2}{\lVert \mathbf{Y}_{ij}-\mathbf{D}\mathbf{X}_{ij} \rVert}_{F}^{2}+\mu \,{\lVert \mathbf{X}_{ij} \rVert}_{2,1}\right\}
\end{equation}
that are simpler to tackle and faster to solve. Here, $\mathbf{Y}_{ij}$ is the $j$'th spectral block of $i$'th spatial group.
\subsection{Selective Spectral Blocks}
\begin{figure}
 \centering
 \includegraphics[width=0.7\textwidth]{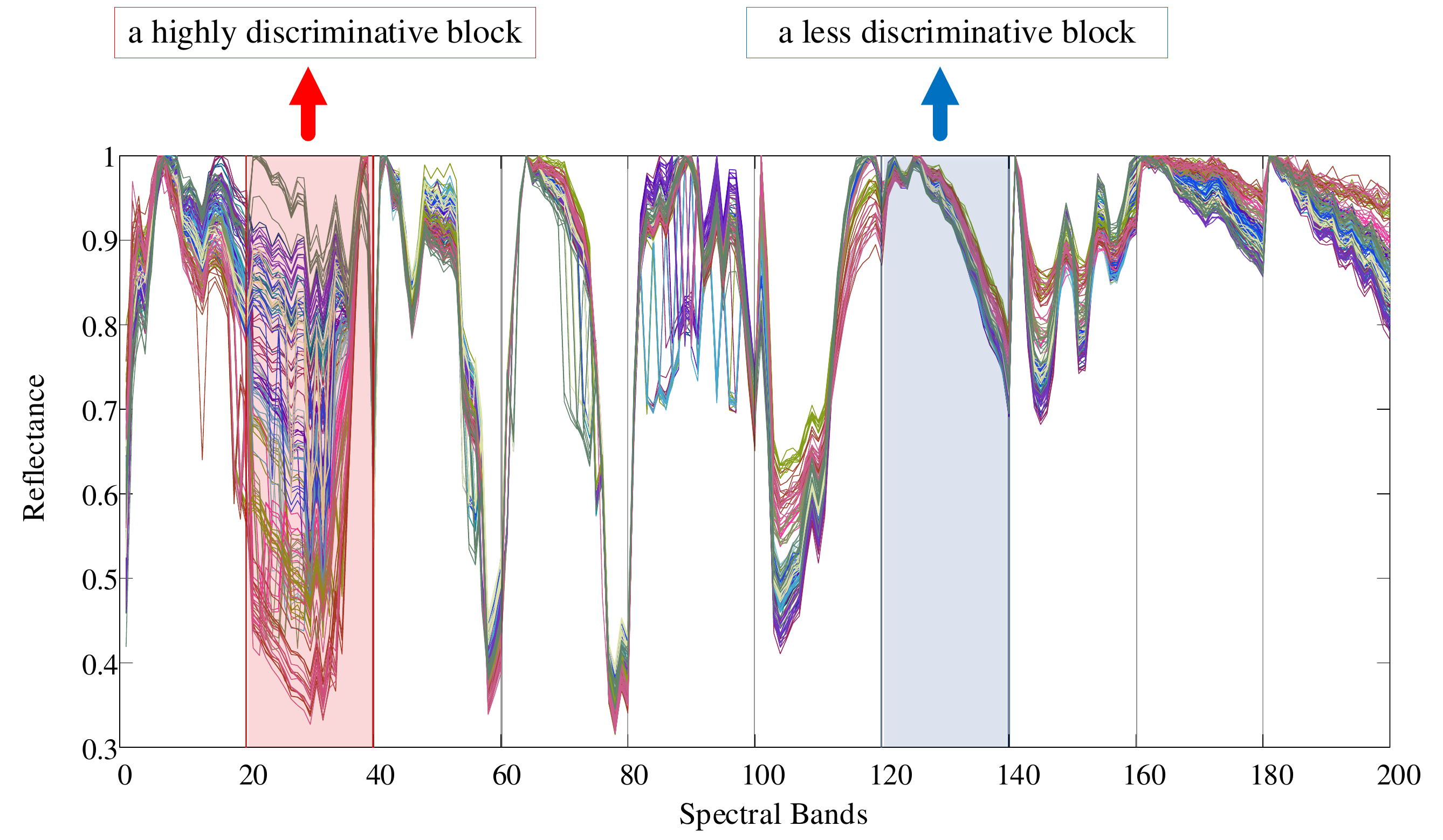}
 \caption{Spectral signatures of 160 pixels of the Indian Pines image. Plots with different colors represent different classes. The red box represents a highly discriminative spectral block while the blue box shows a less discriminative spectral block.}\label{fig.Signatures}
\end{figure}
Spectral redundancy of the HSIs means that not all of the bands take an effective role in distinguishing between classes. Figure 3 depicts the spectral signature of 160 pixels from the Indian Pines image that contains 16 different classes. In this figure, plots with different colors represent different classes. It is observable in the figure that for example, the spectral bands of 20-40 (the red box) are very discriminative while the spectral bands of 120-140 (the blue box) have similar reflectance for different classes.

The proposed sparse structure provides the opportunity to exploit this redundancy. For this end, a criterion can be defined to determine the most discriminative spectral blocks (named as active blocks) and discard the rest (inactive blocks). This enables us to have a multispectral view of the HSIs and drop the blocks with fewer discriminability, and consider the discriminative ones for a further hyperspectral view. To achieve this, instead of the identity matrix ($\mathbf{I}$), a binary diagonal matrix named $\mathbf{W}$ is used. The diagonal entries of $\mathbf{W}$ corresponding to the active blocks are set to 1, and the rest are set to 0.

In this study, the variance of the spectral blocks is used as a criterion to determine the active blocks. That is, for the $j$’th spectral block, the corresponding diagonal entry of $\mathbf{W}$ is defined as:
\begin{equation}\label{Eq.12}
    w_{j}=\mathrm{u}(\sigma^{2}_{j}-T)
\end{equation}
\begin{equation}\label{Eq.13}
    \sigma^{2}_{j}=\mathrm{Var}(\mathbf{m}_{j}).
\end{equation}
Here, $\mathrm{u}$ is the unit function, and $T$ is a threshold. Note that different values of $T$ will result in different number of active blocks. This parameter will be tuned in the next section. $\mathbf{m}_{j}$ is a vector that collects the mean values of the columns of the $j$’th spectral block. Therefore, the optimization problem of \eqref{Eq.10} is written as:
\begin{equation}\label{Eq.14}
 \underset{\mathbf{X}_{i}}{\operatorname{argmin}}\, \, \left\{ \frac{1}{2}{\lVert \mathbf{Y}_{i}-(\mathbf{W}\otimes \mathbf{D})\mathbf{X}_{i} \rVert}_{F}^{2}+\mu \,{\lVert \mathbf{X}_{i} \rVert}_{2,1}\right\}.
\end{equation}
The experimental results presented in the next section verify that inactivating the less effective blocks further reduces the computational burden and the instability of the sparse model resulting in better classification accuracy. \hyperref[Algorithm.1]{Algorithm \ref*{Algorithm.1}} summarizes the proposed SMSB method.

\begin{algorithm}[t]
\caption{Sparse Modeling of Spectral Blocks (SMSB)}
\label{Algorithm.1}
\textbf{Input:} 1) The hyperspectral image $\mathbf{Y}$. 2)Predefined parameters according to \hyperref[table.tune]{Table \ref*{table.tune}}. 3) Labeled train and test data $\mathbf{Y}_{train}$ and $\mathbf{Y}_{test}$ \\
\textbf{Output:} Obtained class labels for $\mathbf{Y}_{test}$ \\
\begin{algorithmic}[1]
\State Segment $\mathbf{Y}$ into spatial groups and spectral blocks $\{\mathbf{Y}_{ij}\}_{i=1,...,G , j=1,...,B}$ (as in \hyperref[fig.hsi]{Figure \ref*{fig.hsi}})
\State Construct $\mathbf{Y}_{s}$ according to \eqref{Eq.8}.
\State Train the dictionary $\mathbf{D}$ according to \eqref{Eq.9}.
\State Extract the set of active spectral blocks and construct $\mathbf{W}$.
\For {$i$'th spatial group}
\State obtain the sparse coefficients $\mathbf{X}_{i}$ according to \eqref{Eq.14}.
\EndFor
\State Train the SVM using the sparse coefficients $\mathbf{X}_{train}$.
\State Obtain the class labels for $\mathbf{X}_{test}$ using the trained SVM.
\end{algorithmic}
\end{algorithm}

\section{Results and Discussion}
\label{section.four}
The following three widely-used datasets were used to evaluate the proposed SMSB method:
\begin{enumerate}
  \item Indian Pines Image: This dataset is taken over the Indian Pines test site in north-western Indiana by the AVIRIS sensor. The main dataset consists of 224 spectral bands in the spectral range of 0.4-2.5 $\mathrm{\mu m}$. In this paper the corrected version of this dataset is used in which 20 water absorption bands (104-108, 150-163, 220) are removed. Each spectral band of the dataset is an image of $145\times 145$ pixels with a spatial resolution of 20 $\sfrac {\mathrm{m}}{\mathrm{pixels}}$.
  \item Pavia University Image: This image is one of the two datasets captured over Pavia, northern Italy by ROSIS sensor. It contains 103 spectral bands in the spectral range of 0.43-0.86 $\mathrm{\mu m}$. Each spectral band is an image of $610\times340$ pixels with a spatial resolution of 1.3 $\sfrac {\mathrm{m}}{\mathrm{pixels}}$ taken over the areas surrounding the University of Pavia, Italy.
  \item Salinas Image: Like the Indian Pines, this image is also taken by AVIRIS sensor and consists 224 spectral bands. 20 water absorption bands (108-112, 154-167, 224) were discarded in these experiments. It covers vegetation and soil areas of Salinas Valley, California. Each spectral band is an image of $512\times 217$ pixels with a spatial resolution of 3.7 $\sfrac {\mathrm{m}}{\mathrm{pixels}}$.
\end{enumerate}

\begin{figure}
        \centering

        \begin{subfigure}{0.33\textheight}
                \centering
                \includegraphics[width=\textwidth]{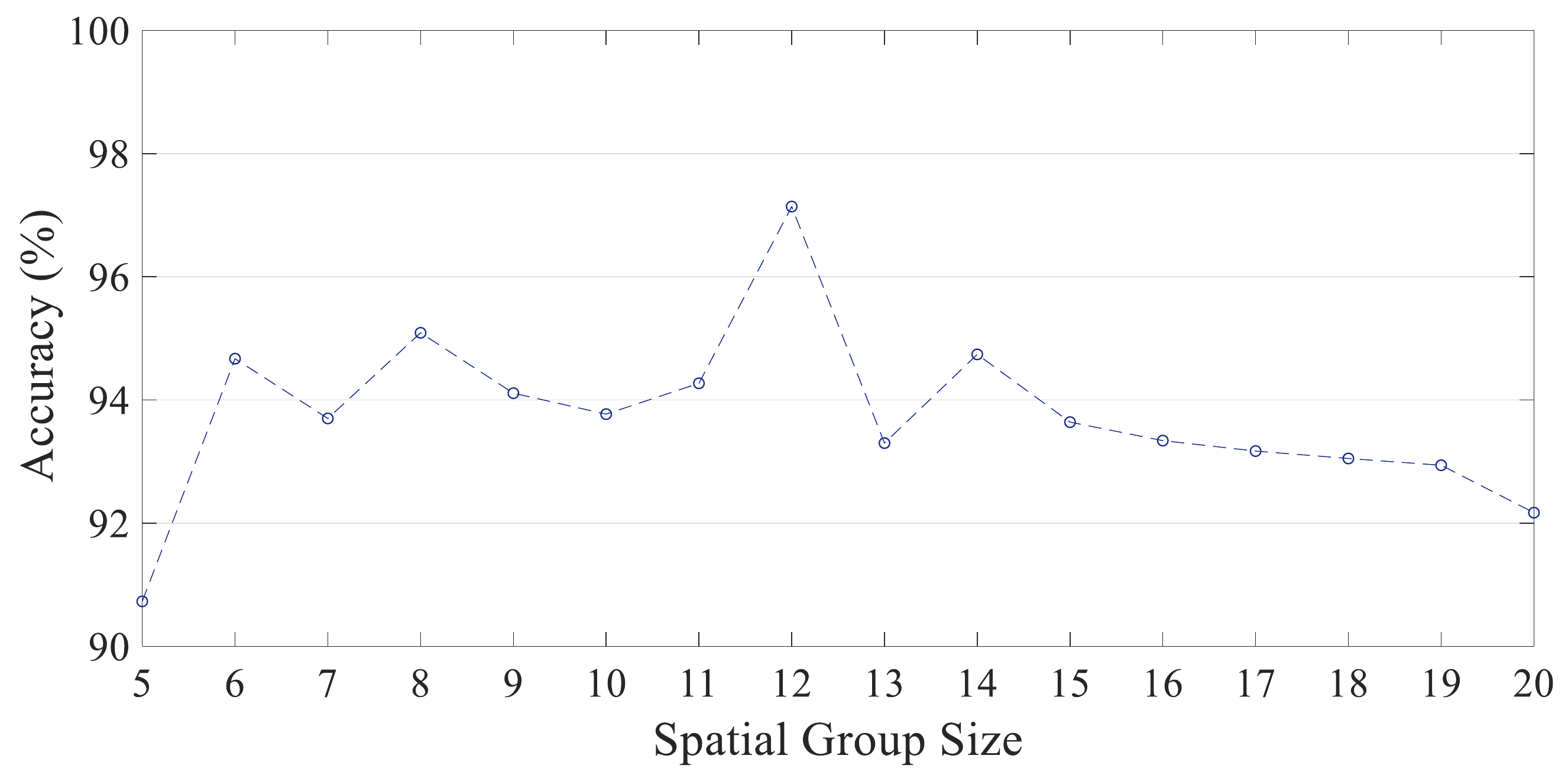}
                \caption{}
                \label{fig.tunea}
        \end{subfigure}
        \begin{subfigure}{0.33\textheight}
                \centering
                \includegraphics[width=\textwidth]{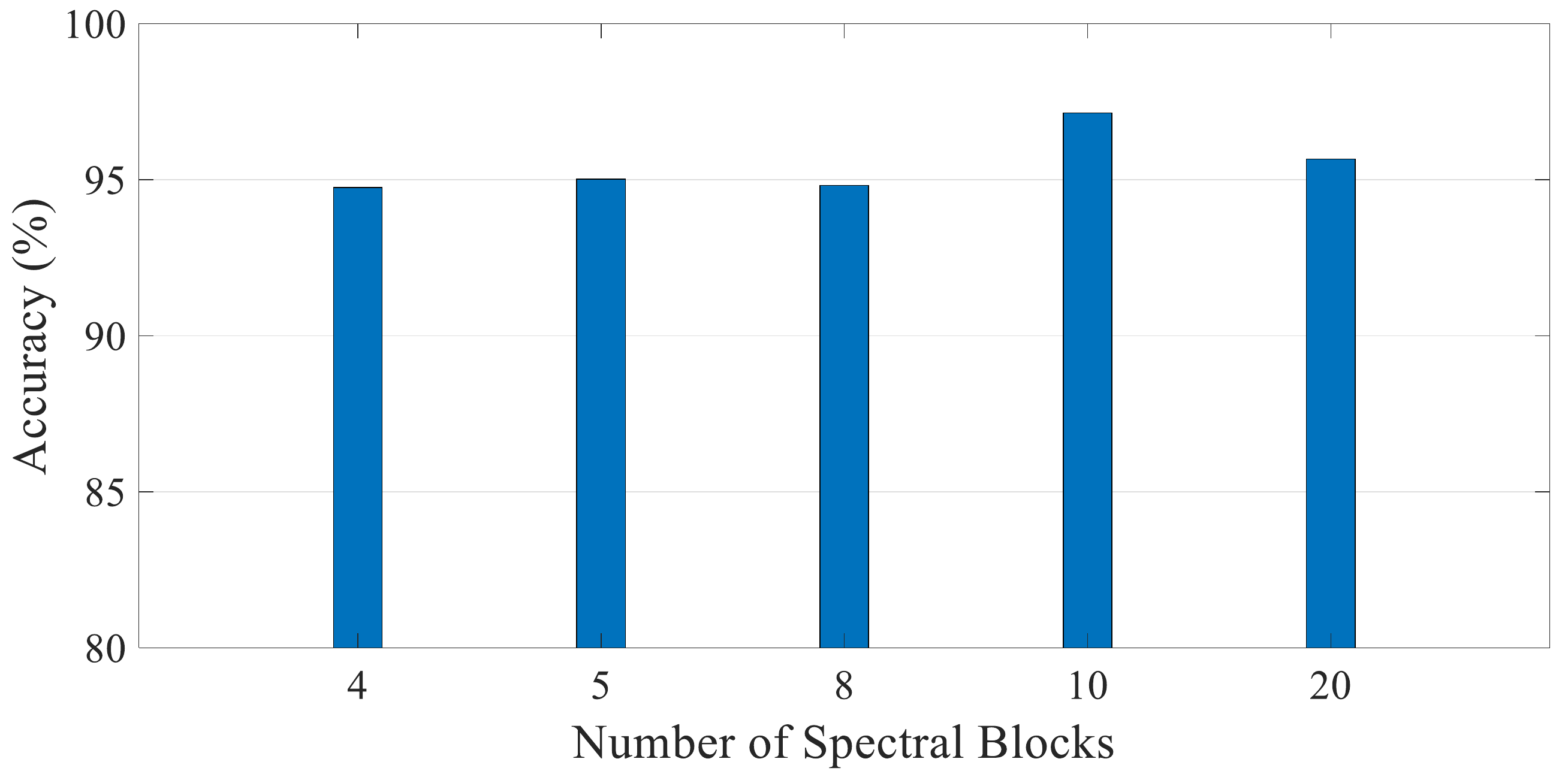}
                \caption{}
                \label{fig.tuneb}
        \end{subfigure}
        \begin{subfigure}{0.33\textheight}
                \centering
                \includegraphics[width=\textwidth]{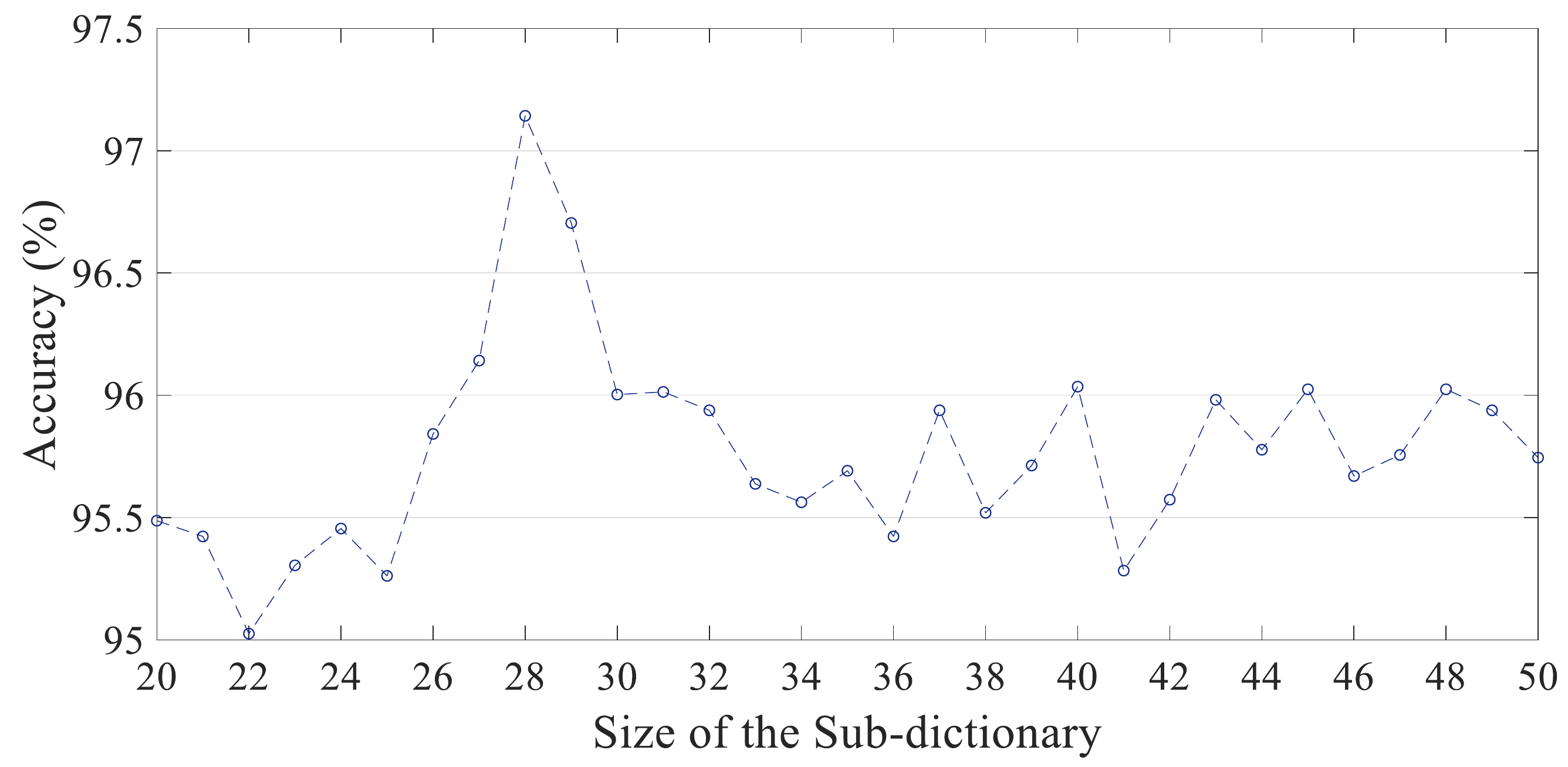}
                \caption{}
                \label{fig.tunec}
        \end{subfigure}
        \begin{subfigure}{0.33\textheight}
                \centering
                \includegraphics[width=\textwidth]{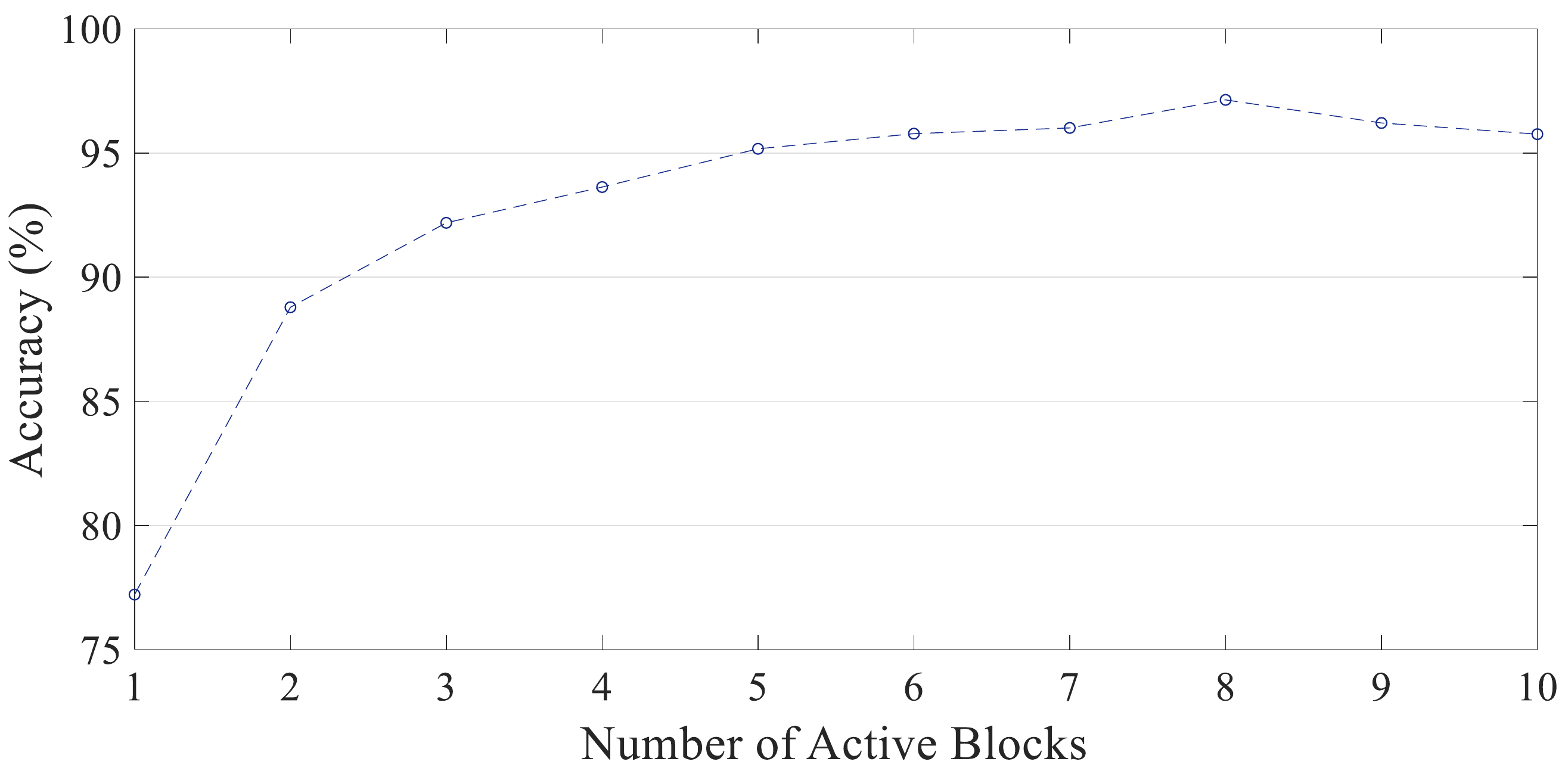}
                \caption{}
                \label{fig.tuned}
        \end{subfigure}
        \caption{The overall classification accuracy of the proposed SMSB method as a function of (a) spatial group size, (b) number of spectral blocks, (c) size of the sub-dictionary and (d) number of active blocks.}
        \label{fig.tune}
\end{figure}
There are four crucial parameters to be tuned before applying the proposed method for the classification of the HSIs. The first parameter is the size of the $m\times m$ spatial groups. \hyperref[fig.tunea]{Figure \ref*{fig.tunea}} shows the accuracy of the system as a function of $m$ for the Indian Pines image. As it can be observed in the figure, the overall accuracy increases as we increase the size of the patches, reaching its highest value for $12\times12$ spatial groups and further increasing the patch size, decreases the accuracy. This can be explained as following: small spatial groups lead to oversampling of the spatial data while large spatial groups lead to undersampling, both resulting in lower classification accuracy. The second parameter is the number of spectral blocks ($B$). \hyperref[fig.tuneb]{Figure \ref*{fig.tuneb}} shows the accuracy obtained with a different number of blocks. Note that, the spectral dimension of the image is segmented into $B$ spectral blocks. Therefore, $B$ is chosen such that the whole number of spectral bands is divisible by $B$ resulting in equal-sized spectral blocks. According to the figure, the best performance was achieved for 10 spectral blocks.

The third parameter is the number of atoms ($k$) of the sub-dictionary. \hyperref[fig.tunec]{Figure \ref*{fig.tunec}} presents the accuracy of the system for different number of atoms. According to the figure, the best performance was achieved for 28 atoms. The last parameter to be tuned is $T$ in \eqref{Eq.8}. Note that changing $T$ results in a different number of active spectral blocks. Therefore, to better illustrate the influence of the selective spectral blocks the accuracy of the system is shown as a function of the number of active blocks in \hyperref[fig.tuned]{Figure \ref*{fig.tuned}}. As expected, the accuracy increases with the number of active blocks. However, for more than 8 active blocks, the accuracy decreases. This indicates that discarding the less effective blocks increases the stability of the sparse model leading to better classification performance.

The same tests were conducted on the Pavia University and Salinas images and the chosen parameters for each dataset is reported in \hyperref[table.tune]{Table \ref*{table.tune}}. It should be noted that the number of spectral blocks are not aliquant to the number of spectral bands in Pavia University and Salinas images (e.g., 103 spectral bands in Pavia University data, while the number of spectral blocks is set as 10). In the current proposed SMSB method, the last 3 bands of the Pavia University image and the last 4 bands of the Salinas image are simply eliminated. However, for future studies more advanced methods can be used to extract smarter spectral blocks and solve this problem.
\begin{table}
  \centering
  \caption{Parameters of the SMSB method for the HSIs used in the experiments.}\label{table.tune}
\begin{tabular}{|c|c c c|}
  \hline
  Parameters & Indian Pines  & Pavia University & Salinas \\
  \hline
  Spatial Group Size ($m\times m$)& $12\times12$ & $13\times 13$ & $32\times 32$\\
  Number of Spectral Blocks ($B$) & 10  & 10 & 7\\
  Number of Sub-dictionary Atoms ($k$) & 28  & 40 & 29 \\
  Number of Active Blocks& 8 & 8 & 7 \\
  \hline
\end{tabular}
\end{table}
\subsection{Classification Results}
The proposed SMSB method is used for classification of the introduced datasets, and its performance is compared with six state-of-the-art methods. The first method, applies SVM \cite{melgani04} with RBF kernel function to the spectral data and its parameters are selected using five-fold cross-validation. The rest of the methods used for comparison are sparse representation-based methods which include SOMP \cite{chen11}, SADL \cite{soltani15}, LGIDL \cite{he16}, CODL \cite{fu18} and LAJSR \cite{peng2019}. SOMP uses training data as dictionary and solves the sparse representation problem in a greedy manner. LAJSR uses square overlapping spatial patches and defines a dictionary constructed from training data and their neighbors. SADL and CODL utilize square non-overlapping spatial groups and employ dictionary learning to improve the representation of the data.  LGIDL uses super-pixel segmentation to define spatial groups and trains a dictionary. For SOMP, the whole training data was used as dictionary and $p$ in $\ell_{p}$-norm was set to $\infty$. The rest of the parameters of the SOMP and all of the parameters of SADL, LGIDL, CODL and LAJSR were chosen according to their corresponding papers. Note that all these methods (except for the LAJSR method) use the obtained sparse coefficients as extracted features and feed them to SVM classifier for final classification. Hence, the same SVM structure is used in all these methods and it is implemented using the LIBSVM toolbox \cite{CC01a}.

Since our method is similar to the SADL and CODL methods except for the utilization of the spectral blocks, our main concentrate was to analyze the effectiveness of the spectral blocks in reducing the processing time while achieving good classification performance. Hence, we have used the same non-overlapping patches as the SADL and CODL methods to have a fair comparison. One of the drawbacks of non-overlapping spatial patches is that they can lead to edge errors. For future studies, one solution to overcome this problem is to use overlapping patches. Another solution is to exploit superpixel segmentation methods to have more adaptive spatial groups. To compare the results, three widely-used measures, i.e. Overall Accuracy (OA), Average Accuracy (AA) and $\kappa$ coefficient, are used.

\begin{figure}
        \centering

        \begin{subfigure}{0.125\textheight}
                \centering
                \includegraphics[width=\textwidth]{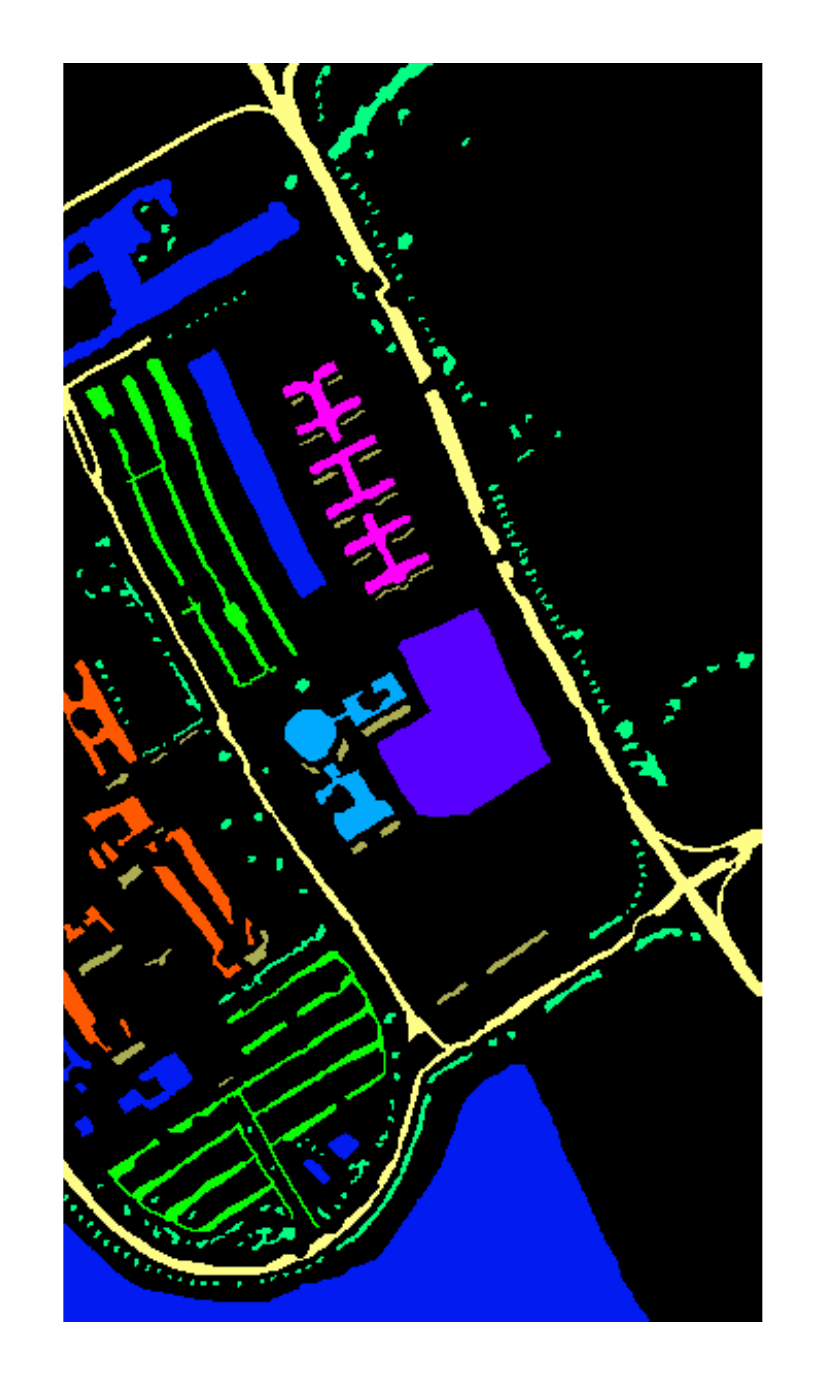}
                \caption{}
                \label{fig.pavia}
        \end{subfigure}
        \begin{subfigure}{0.125\textheight}
                \centering
                \includegraphics[width=\textwidth]{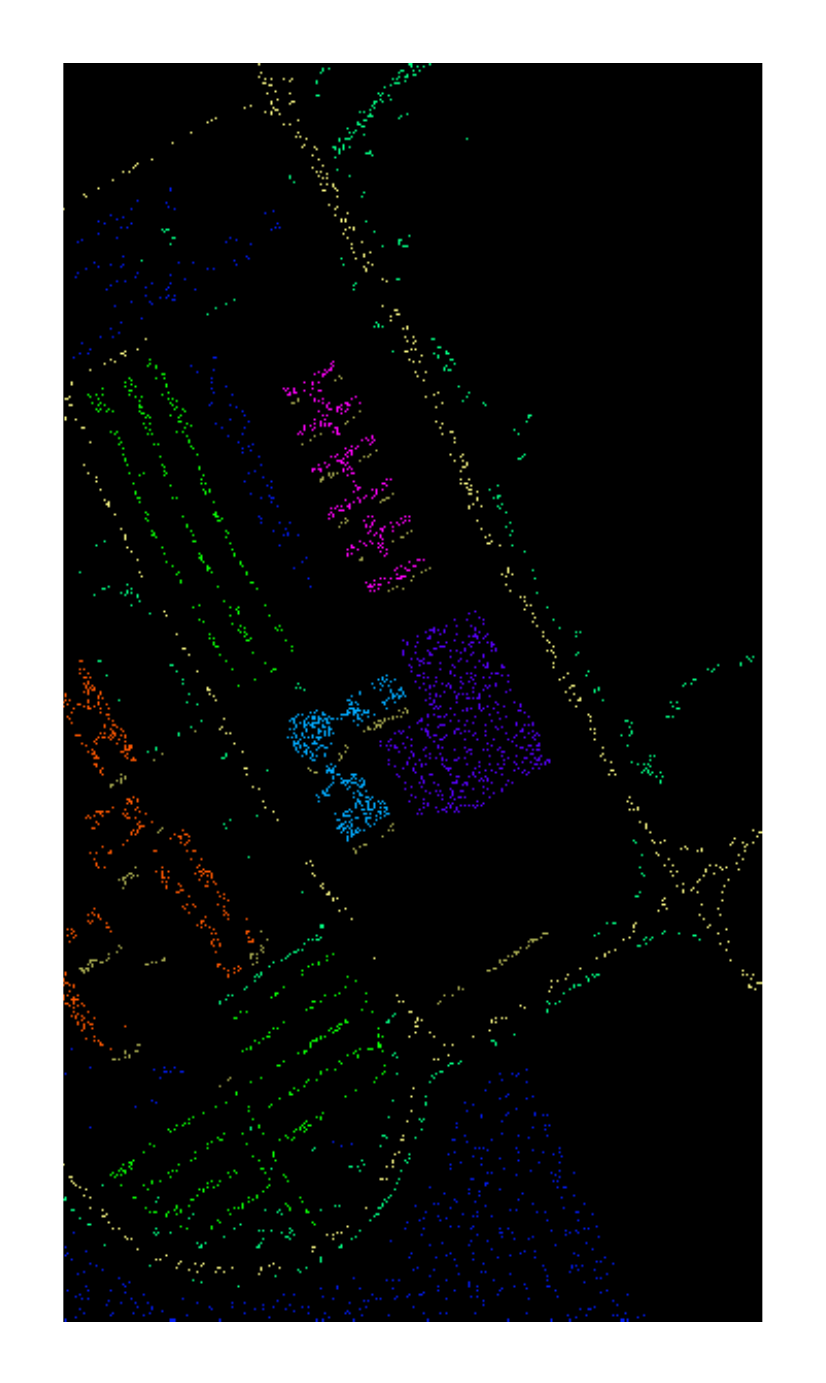}
                \caption{}
                \label{fig.paviatrain}
        \end{subfigure}
        \begin{subfigure}{0.125\textheight}
                \centering
                \includegraphics[width=\textwidth]{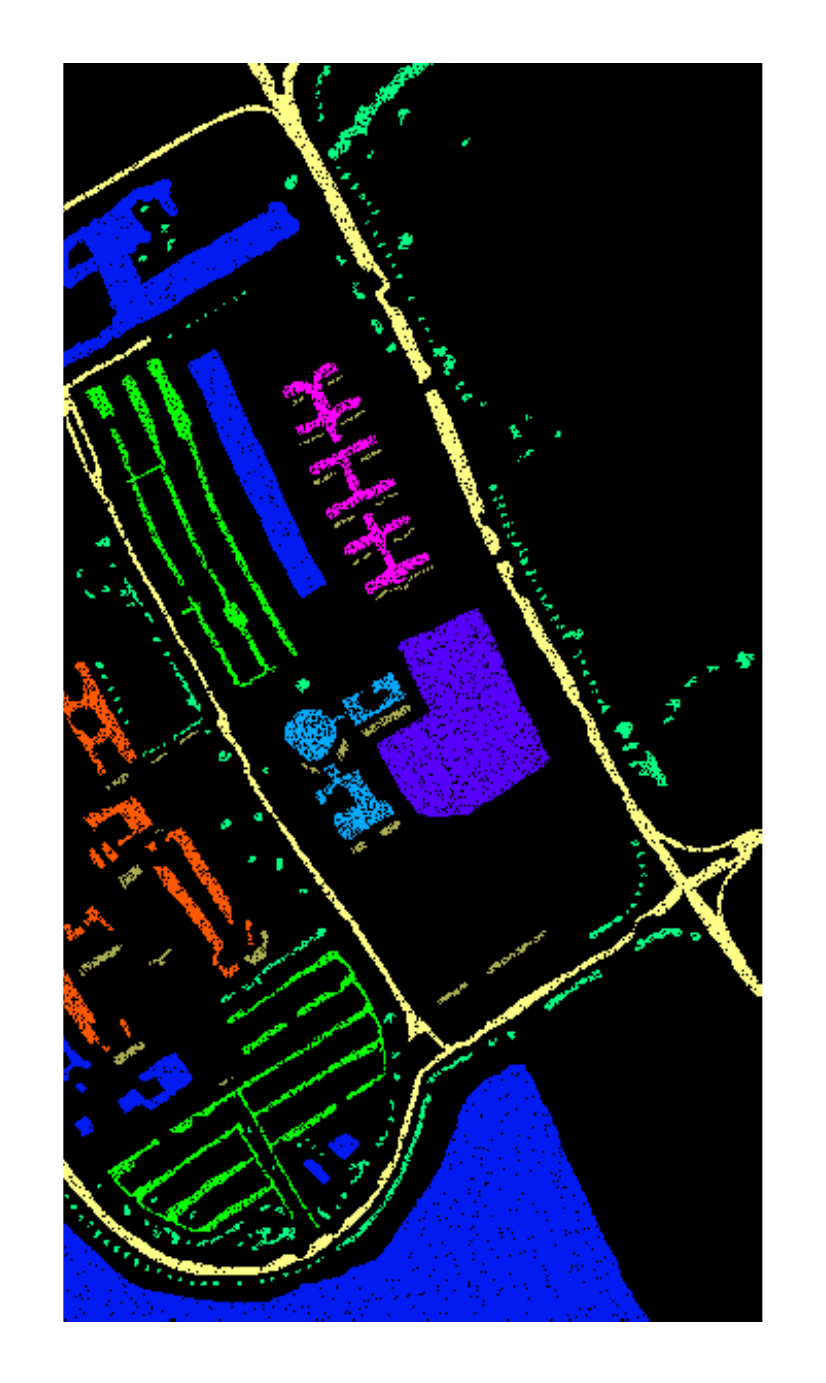}
                \caption{}
                \label{fig.paviatest}
        \end{subfigure}
        \begin{subfigure}{0.125\textheight}
                \centering
                \includegraphics[width=\textwidth]{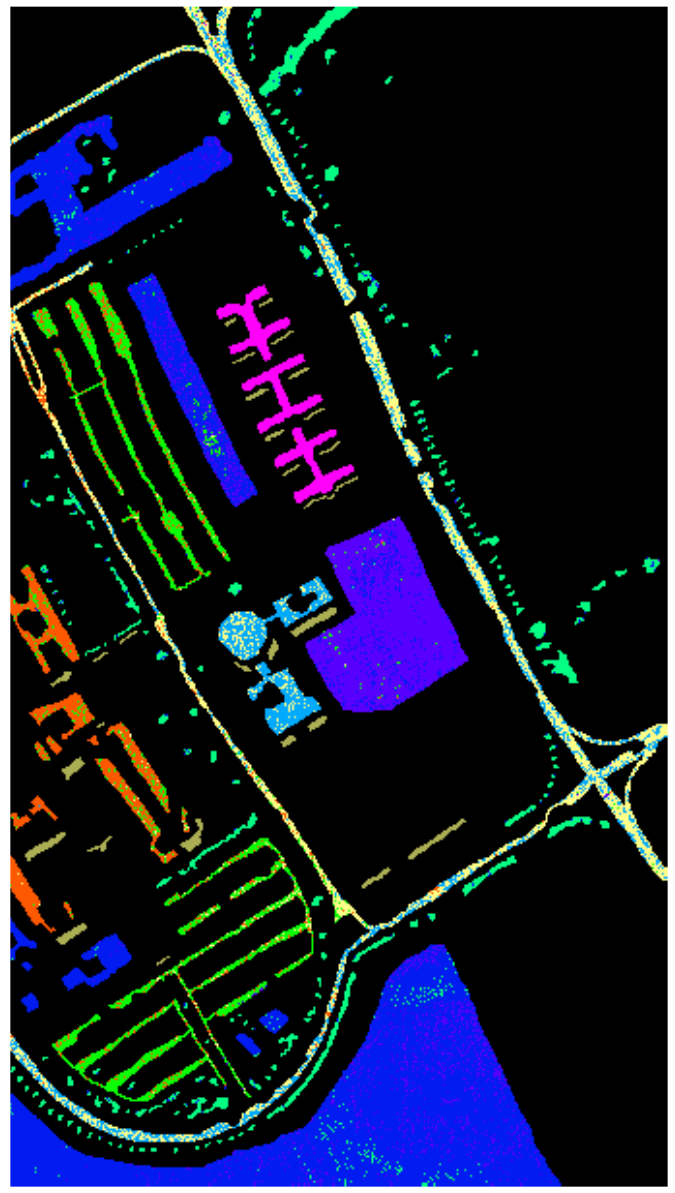}
                \caption{}
                \label{fig.paviaSVM}
        \end{subfigure}
        \begin{subfigure}{0.125\textheight}
                \centering
                \includegraphics[width=\textwidth]{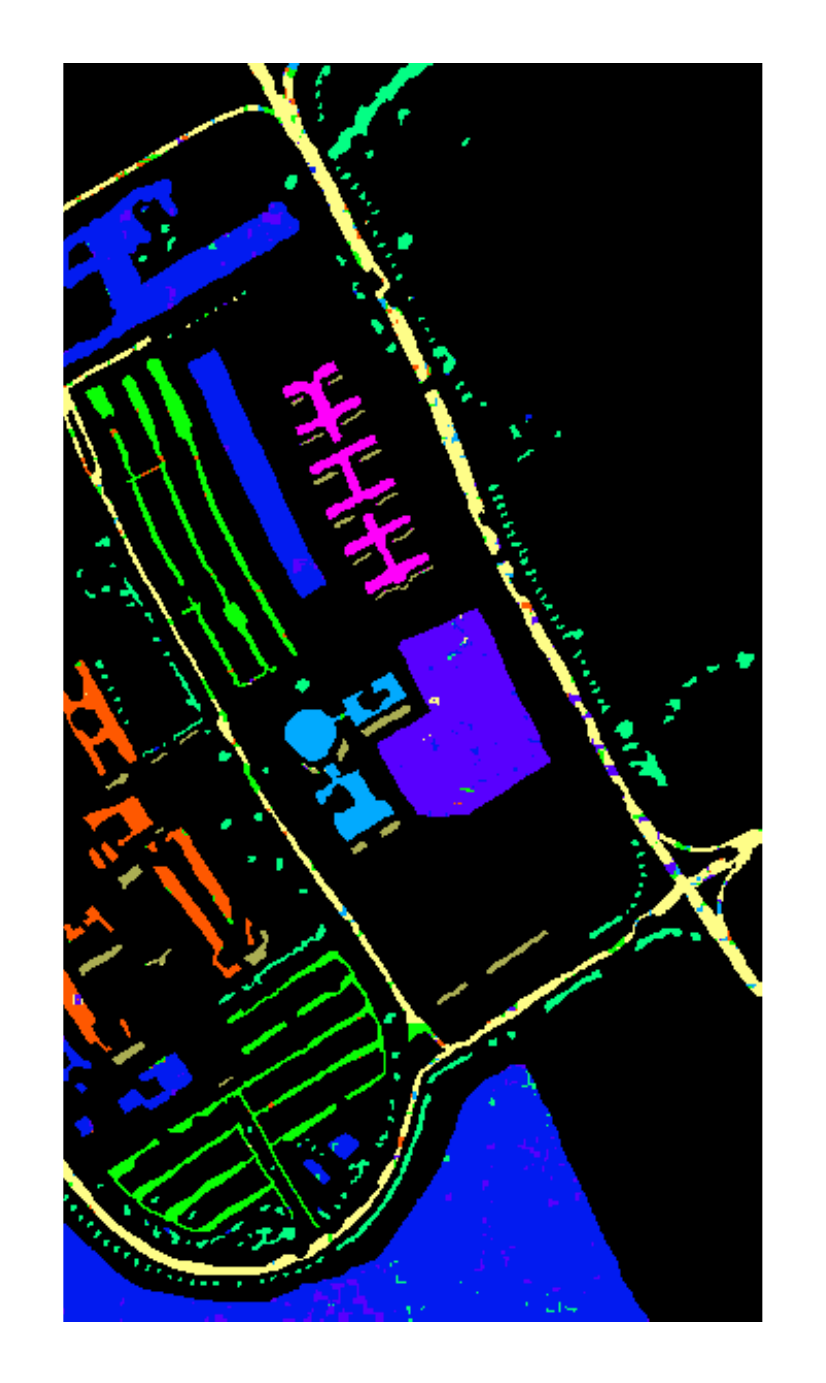}
                \caption{}
                \label{fig.paviaSOMP}
        \end{subfigure}
        \begin{subfigure}{0.125\textheight}
                \centering
                \includegraphics[width=\textwidth]{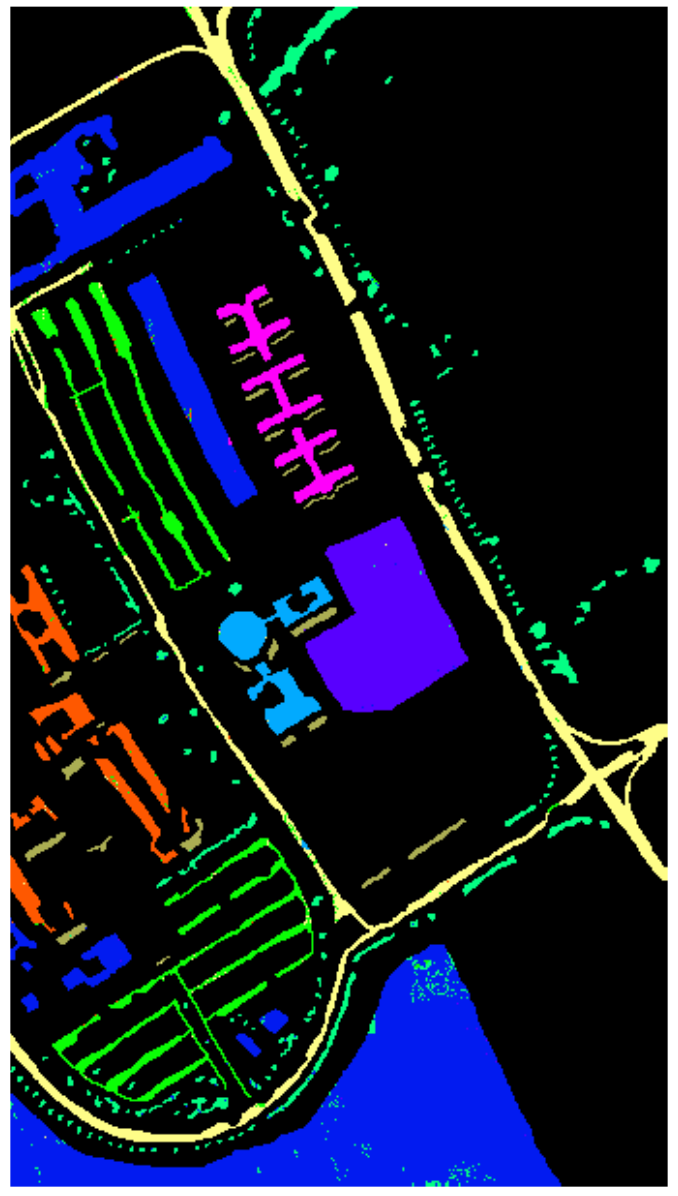}
                \caption{}
                \label{fig.paviaSADL}
        \end{subfigure}
        \begin{subfigure}{0.125\textheight}
                \centering
                \includegraphics[width=\textwidth]{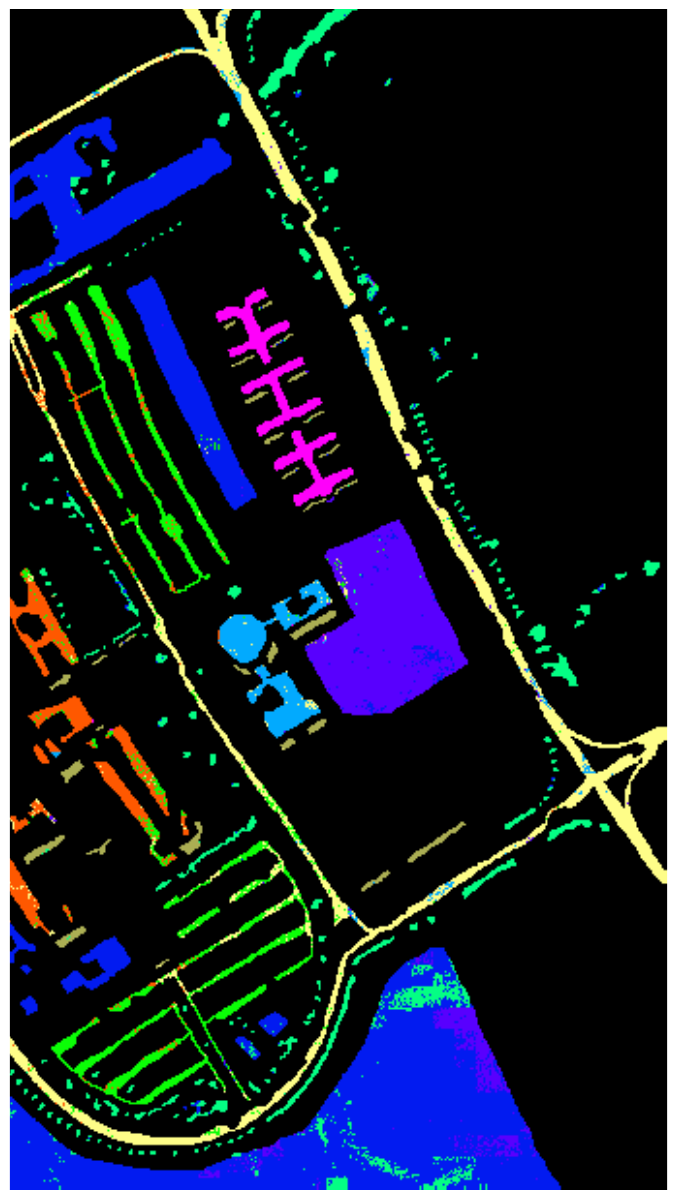}
                \caption{}
                \label{fig.paviaLGIDL}
        \end{subfigure}
        \begin{subfigure}{0.125\textheight}
                \centering
                \includegraphics[width=\textwidth]{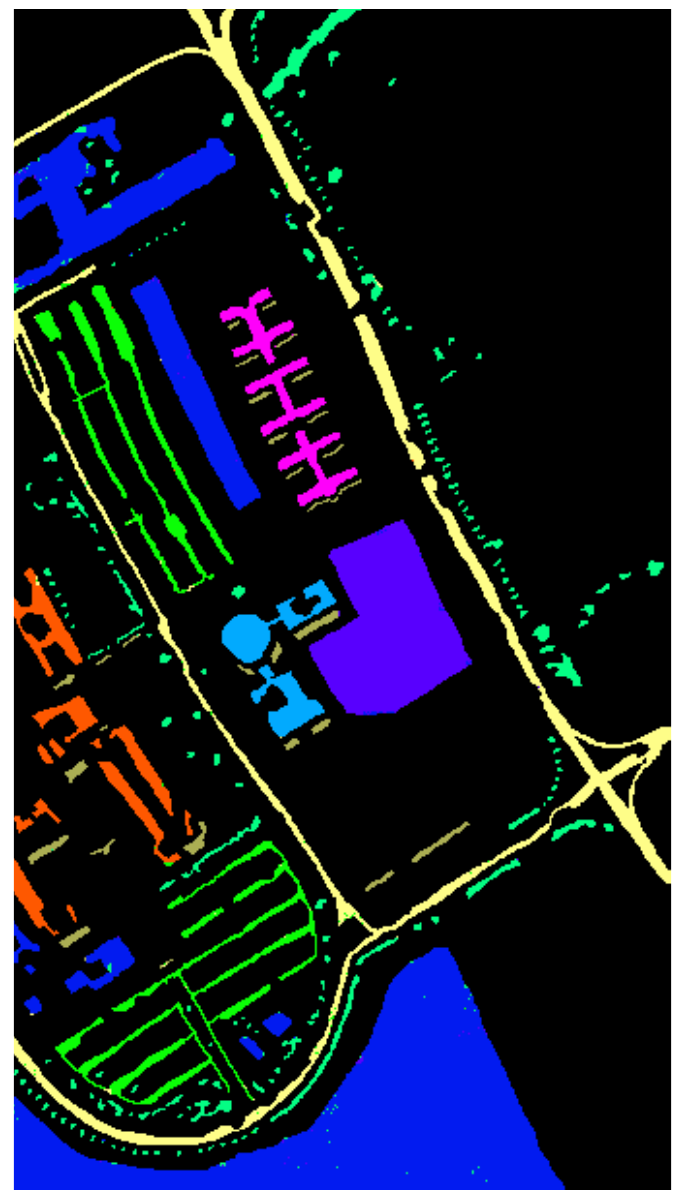}
                \caption{}
                \label{fig.paviaCODL}
        \end{subfigure}
        \begin{subfigure}{0.125\textheight}
                \centering
                \includegraphics[width=\textwidth]{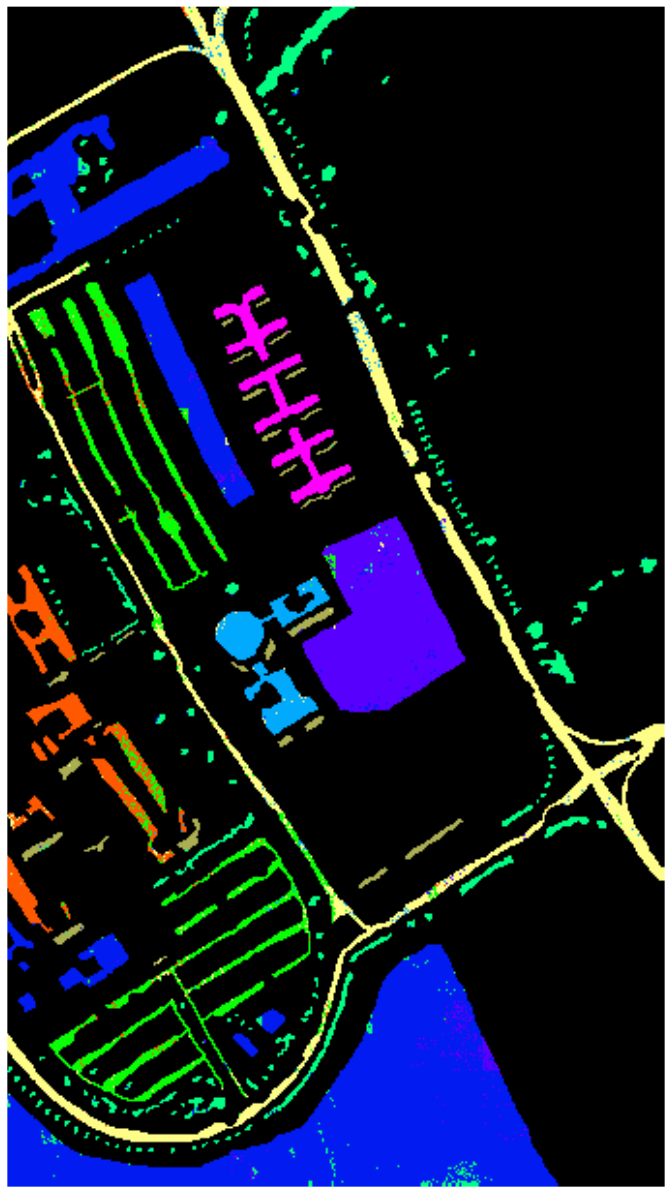}
                \caption{}
                \label{fig.paviaLAJSR}
        \end{subfigure}
                \begin{subfigure}{0.125\textheight}
                \centering
                \includegraphics[width=\textwidth]{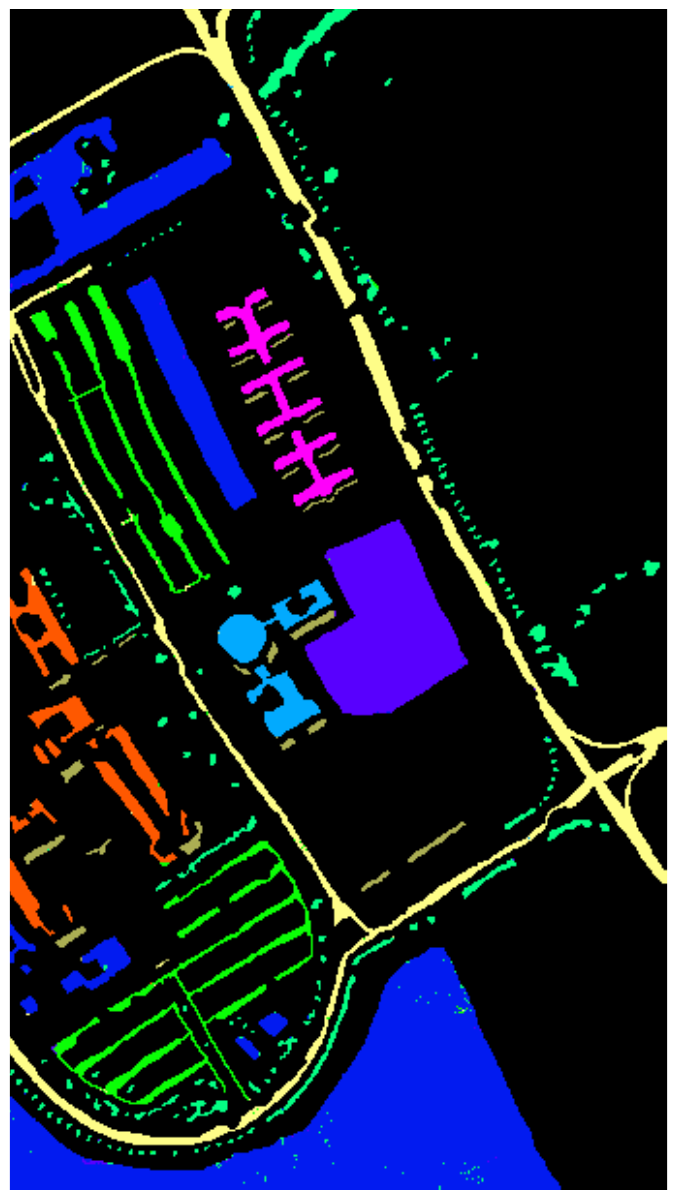}
                \caption{}
                \label{fig.paviaSMSB}
        \end{subfigure}
        \caption{Classification results for Pavia University image. (a) Ground truth. (b) Training data. (c) Test data. (d) SVM \cite{melgani04}. (e) SOMP \cite{chen11}. (f) SADL \cite{soltani15}. (g) LGIDL \cite{he16}. (h) CODL \cite{fu18}. (i) LAJSR \cite{peng2019}. (j) SMSB.}
        \label{fig.paviaresult}
\end{figure}
\begin{table}
  \centering
\caption{Detailed classification results including class-specific accuracy (\%), overall accuracy (OA), average accuracy (AA), $\kappa$ coefficient and processing time (sec) for Pavia University image with different classification approaches.}
\resizebox{0.92\textwidth}{0.15\textheight}{%
\label{table.Pavia}
\begin{tabular}{|c|c|c c|c |c |c |c |c |c |c|}
  \hline
  Class & Name & Train & Test & SVM & SOMP & SADL & LGIDL & CODL & LAJSR & SMSB \\
  \hline \hline
  {\cellcolor [rgb]{0.9961,0.9922,0.5352}1} & Asphalt & 548 & 6083 & 59.71 \!$\pm 1.87$& 85.76 \!$\pm 0.79$& 96.55 \!$\pm 0.83$& 89.84 \!$\pm 1.21$& 98.93 \!$\pm 0.48$& 90.89 \!$\pm 0.77$&\textbf{99.11} \!$\pm 0.11$\\
  {\cellcolor [rgb]{0.0117,0.1094,0.9414}2} & Meadows & 540 & 18109 & 88.76 \!$\pm 0.95$& 95.78 \!$\pm 0.81$& 98.13 \!$\pm 0.51$& 97.31 \!$\pm 0.71$& 98.95 \!$\pm 0.78$& 96.55 \!$\pm 0.12$& \textbf{98.97} \!$\pm 0.41$\\
  {\cellcolor [rgb]{0.9961,0.3477,0.0039}3} & Gravel & 392 & 1707 & 82.02 \!$\pm 1.02$& 98.83 \!$\pm 0.26$& 98.30 \!$\pm 0.64$&  89.16 \!$\pm 0.93$& \textbf{98.95} \!$\pm 0.57$&86.76 \!$\pm 1.21$& 98.89 \!$\pm 0.36$\\
  {\cellcolor [rgb]{0.0195,0.9961,0.5195}4} & Trees & 524 & 2540 & 94.02 \!$\pm 0.11$& 97.67 \!$\pm 0.91$& 98.11 \!$\pm 0.87$& 97.91 \!$\pm 0.41$& \textbf{99.13} \!$\pm 0.44$& 97.79 \!$\pm 0.73$& 98.74 \!$\pm 0.87$\\
  {\cellcolor [rgb]{0.9961,0.0078,0.9805}5} & Painted metal sheets & 265 & 1080 & 99.72 \!$\pm 0.21$& 99.99 \!$\pm 0.08$& \textbf{100} \!$\pm 0$& \textbf{100} \!$\pm 0$& \textbf{100} \!$\pm 0$&\textbf{100} \!$\pm 0$& \textbf{100} \!$\pm 0$\\
  {\cellcolor [rgb]{0.3477,0.0039,0.9961}6} & Bare Soil & 532 & 4497 & 88.21 \!$\pm 0.87$& 97.03 \!$\pm 0.23$& 99.73 \!$\pm 0.23$& 98.53 \!$\pm 0.24$& 99.29 \!$\pm 0.15$& 97.37 \!$\pm 0.33$&\textbf{99.87} \!$\pm 0.17$\\
  {\cellcolor [rgb]{0.0117,0.6680,0.9961}7} & Bitumen & 375 & 955 & 69.11 \!$\pm 0.98$& 99.60 \!$\pm 0.03$& 99.37 \!$\pm 0.43$& 97.91 \!$\pm 0.49$& \textbf{99.90} \!$\pm 0.07$& 97.17 \!$\pm 0.79$& 99.79 \!$\pm 0.22$\\
  {\cellcolor [rgb]{0.0469,0.9961,0.0273}8} & Self-Blocking Bricks & 514 & 3168 & 74.72 \!$\pm 0.47$& 97.52 \!$\pm 0.81$& 97.76 \!$\pm 0.91$& 92.83 \!$\pm 0.92$& \textbf{99.12} \!$\pm 0.09$& 91.41 \!$\pm 0.69$&98.99 \!$\pm 0.61$ \\
  {\cellcolor [rgb]{0.6719,0.6836,0.3281}9} & Shadows & 231 & 716 & \textbf{99.82} \!$\pm 0.10$& 96.52 \!$\pm 0.47$& 98.60 \!$\pm 1.02$& 98.31 \!$\pm 0.76$& 98.32 \!$\pm 0.14$& 96.65 \!$\pm 1.12$&98.04 \!$\pm 0.72$\\
  \hline \hline
  OA& - & - & - & 83.08 \!$\pm 1.31$& 94.98 \!$\pm 0.83$& 98.13 \!$\pm 0.75$&  95.70 \!$\pm 0.93$& 99.05 \!$\pm 0.50$& 95.10 \!$\pm 0.76$&\textbf{99.11} \!$\pm 0.21$\\
  \hline
  AA& - & - & - & 84.01 \!$\pm 0.82$& 96.52 \!$\pm 0.25$& 98.51 \!$\pm 0.31$& 95.76 \!$\pm 0.55$& \textbf{99.18} \!$\pm 0.23$&94.96 \!$\pm 0.44$& 99.16 \!$\pm 0.13$\\
  \hline
  $\kappa$& - & - & - & 77.41 \!$\pm 1.04$& 93.20 \!$\pm 0.71$& 97.46 \!$\pm 0.63$& 94.16 \!$\pm 0.84$& 98.71 \!$\pm 0.41$ &93.35 \!$\pm 0.65$& \textbf{98.79} \!$\pm 0.18$\\
  \hline \hline
  Time (sec) & - & - & - & 40 \!$\pm 1$& 137 \!$\pm 4$& 313 \!$\pm 6$& 391 \!$\pm 5$& 154 \!$\pm 4$& 432 \!$\pm 8$& \textbf{61} \!$\pm 2$\\
  \hline
\end{tabular}
}
\end{table}
\textit{Pavia University Image}: \hyperref[fig.pavia]{Figure \ref*{fig.pavia}} presents the ground truth image of the Pavia University dataset which contains 9 different classes, and the colors are specified in \hyperref[table.Pavia]{Table \ref*{table.Pavia}}. The exact numbers of train and test data for each class are presented in \hyperref[table.Pavia]{Table \ref*{table.Pavia}} and visualized in \hyperref[fig.paviaresult]{Figure \ref*{fig.paviaresult}}. The proposed SMSB method, along with six other state-of-the-art methods, are used to classify this dataset. The detailed classification results for different methods, including the accuracies obtained for each class, overall accuracy, average accuracy, kappa coefficient and the processing time are presented in \hyperref[table.Pavia]{Table \ref*{table.Pavia}}. The reported values are the mean and standard deviation values obtained from ten runs. Furthermore, the visual classification results are depicted in \hyperref[fig.paviaresult]{Figure \ref*{fig.paviaresult}}.

According to \hyperref[table.Pavia]{Table \ref*{table.Pavia}}, all of the methods outperform SVM due to the additional spatial information and sparse representation. Among these methods SADL, LGIDL, CODL and the proposed SMSB method are dictionary learning based methods and benefited from the trained dictionary, they improve the classification performance in compared with SOMP and LAJSR methods which use training data as dictionary. Considering the utilized spatial groups, SADL, CODL and SMSB use similar square spatial patches and outperform the LGIDL and LAJSR methods which use super-pixel and overlapping square spatial groups, respectively.

Further observations can be made by comparing the proposed SMSB method with other approaches. The SMSB differs from SADL and CODL in the use of spectral blocks. In SMSB, exploiting the spectral blocks increases the overall accuracy about 0.98\% compared to the SADL method, and as it can be seen in \hyperref[table.Pavia]{Table \ref*{table.Pavia}} this improvement is achieved in much less processing time. This improvement can also be seen by comparing \hyperref[fig.paviaSADL]{Figure \ref*{fig.paviaSADL}} and \hyperref[fig.paviaSMSB]{Figure \ref*{fig.paviaSMSB}}. Take Meadows and Gravel classes as examples where the SMSB method has notably decreased the intra-region errors. Therefore, comparing the proposed method with SADL and CODL methods verify the effectiveness of spectral blocks in stabilizing the sparse coefficients and reducing the computational burden. Moreover, \hyperref[table.Pavia]{Table \ref*{table.Pavia}} suggests that the CODL and SMSB perform better than the rest of the methods for classifying the Pavia University dataset with SMSB achieving 0.06\% higher overall accuracy in 93 sec less processing time.

\begin{figure}
        \centering

        \begin{subfigure}{0.125\textheight}
                \centering
                \includegraphics[width=\textwidth]{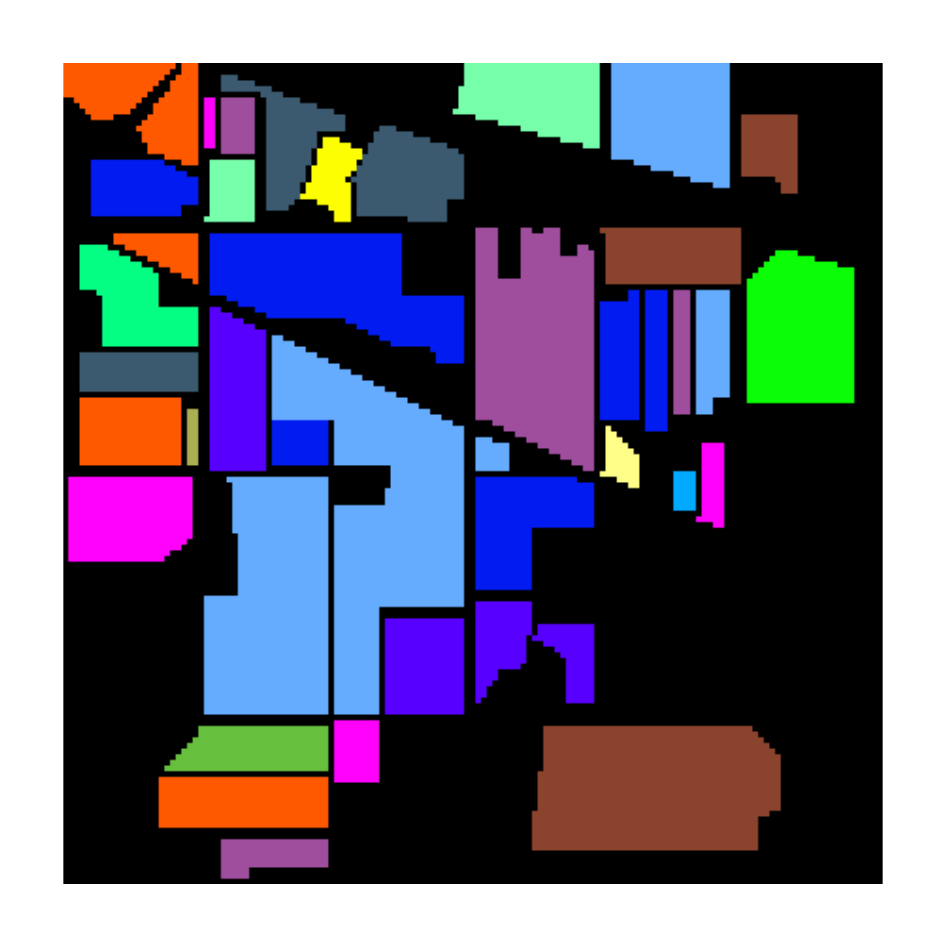}
                \caption{}
                \label{fig.indian}
        \end{subfigure}
        \begin{subfigure}{0.125\textheight}
                \centering
                \includegraphics[width=\textwidth]{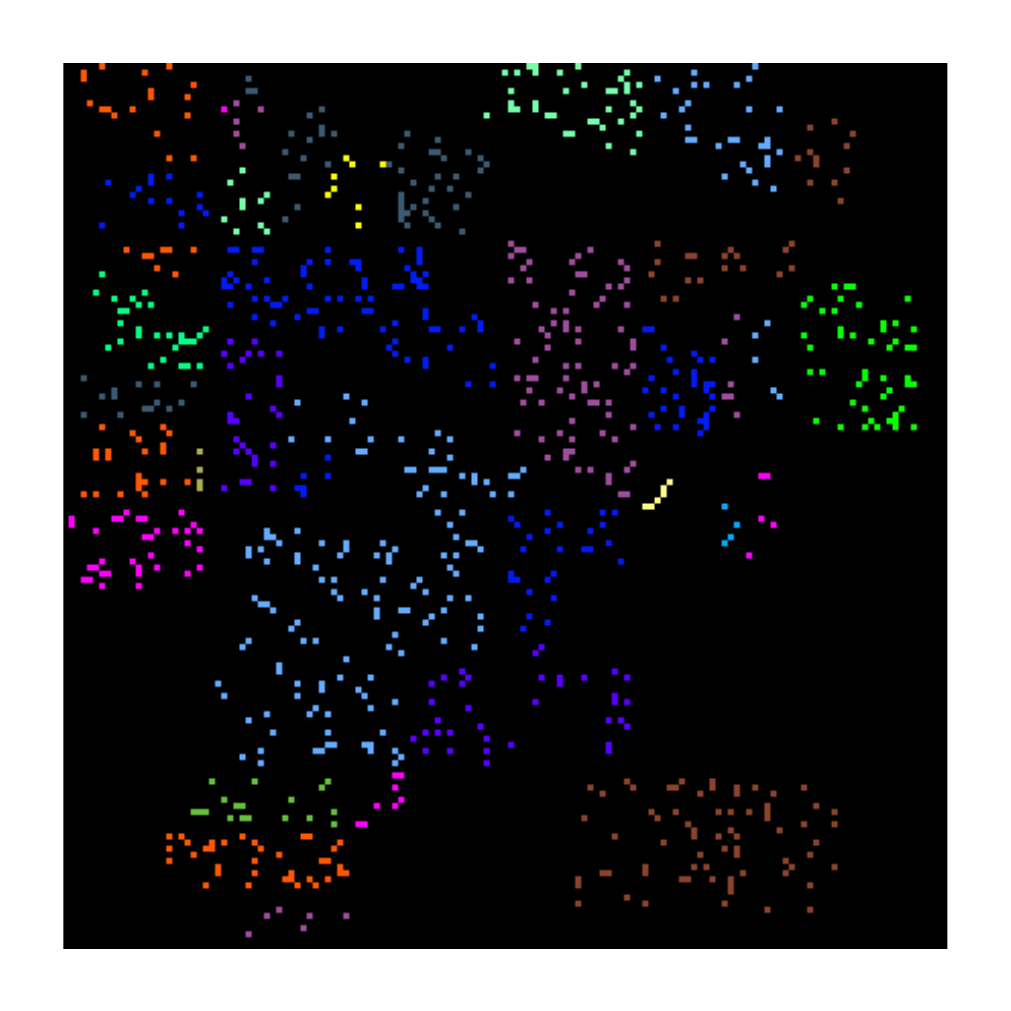}
                \caption{}
                \label{fig.indiantrain}
        \end{subfigure}
        \begin{subfigure}{0.125\textheight}
                \centering
                \includegraphics[width=\textwidth]{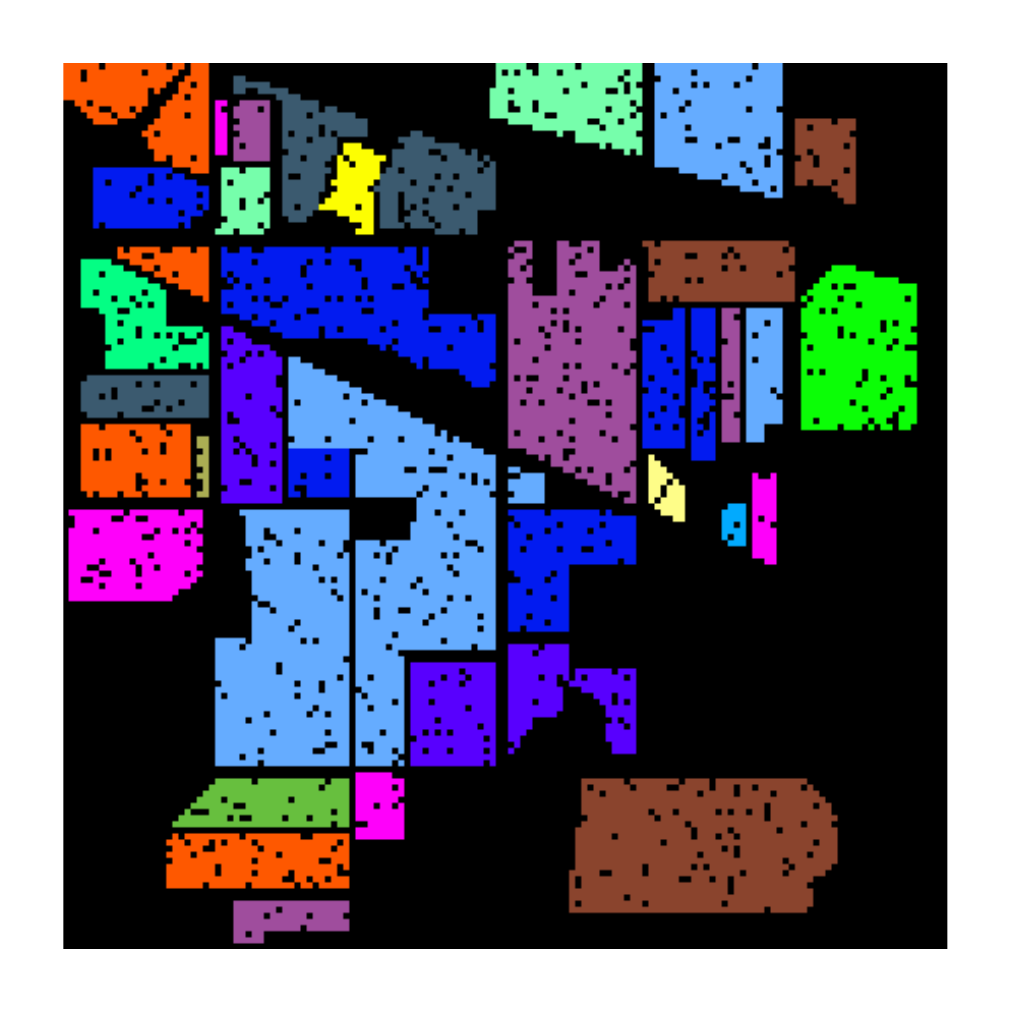}
                \caption{}
                \label{fig.indiantest}
        \end{subfigure}
        \begin{subfigure}{0.125\textheight}
                \centering
                \includegraphics[width=\textwidth]{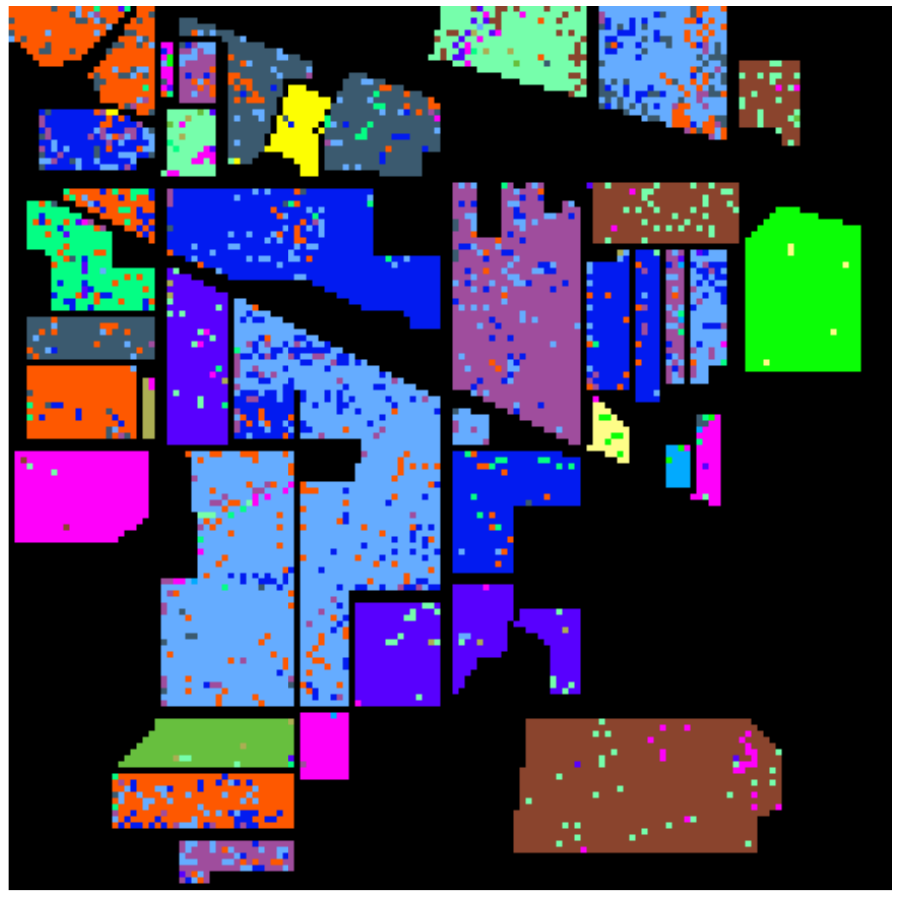}
                \caption{}
                \label{fig.indianSVM}
        \end{subfigure}
        \begin{subfigure}{0.125\textheight}
                \centering
                \includegraphics[width=\textwidth]{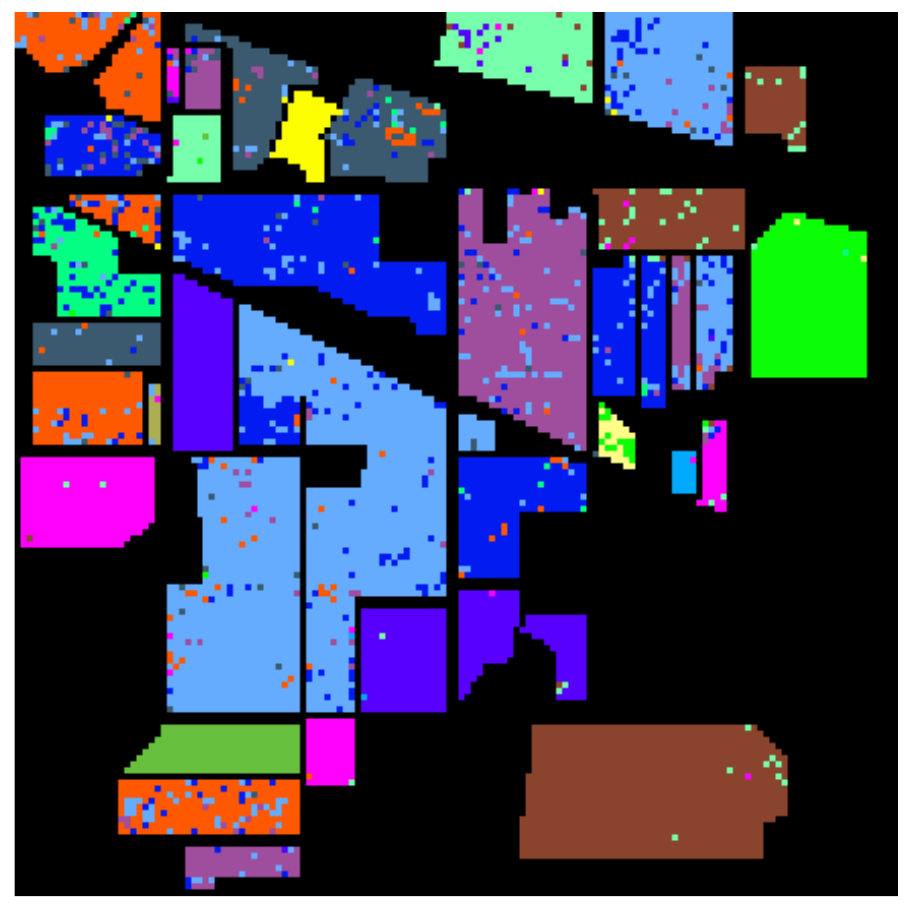}
                \caption{}
                \label{fig.indianSOMP}
        \end{subfigure}
        \begin{subfigure}{0.125\textheight}
                \centering
                \includegraphics[width=\textwidth]{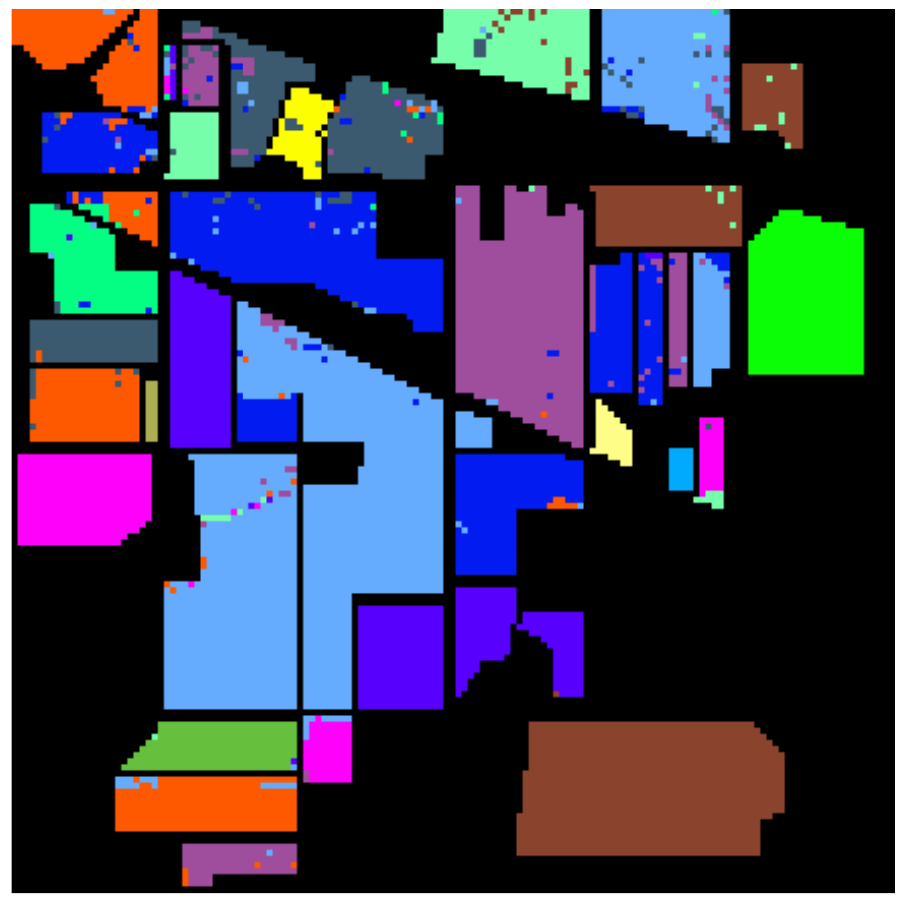}
                \caption{}
                \label{fig.indianSADL}
        \end{subfigure}
        \begin{subfigure}{0.125\textheight}
                \centering
                \includegraphics[width=\textwidth]{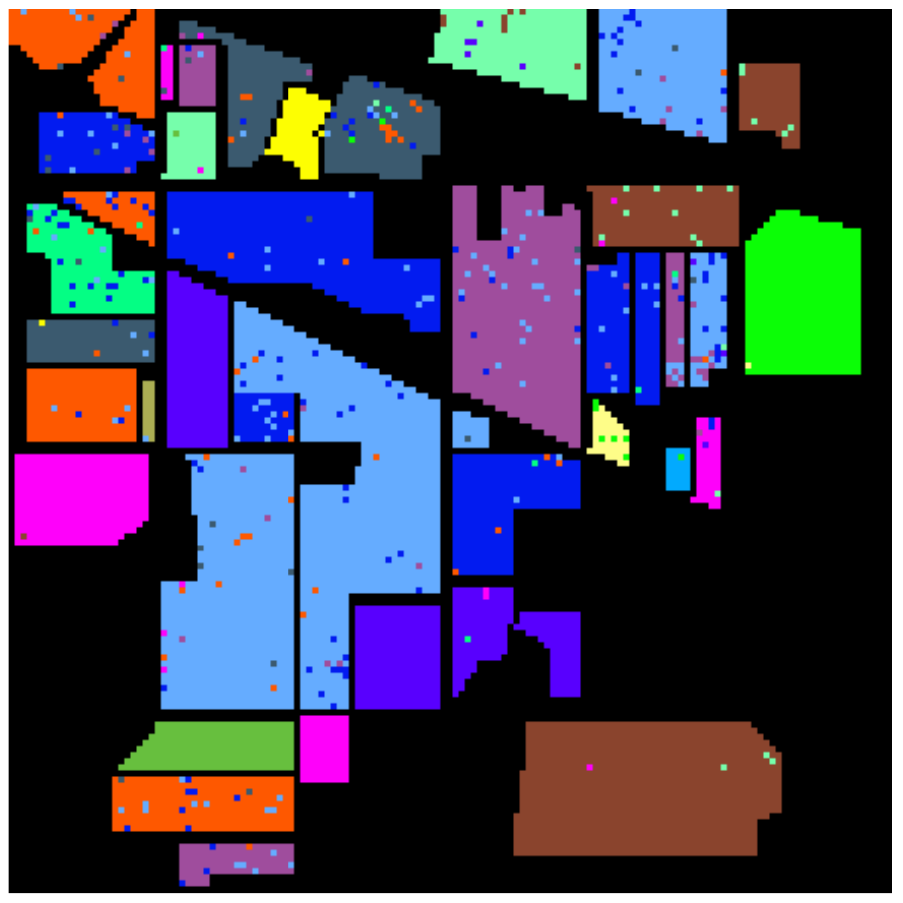}
                \caption{}
                \label{fig.indianLGIDL}
        \end{subfigure}
        \begin{subfigure}{0.125\textheight}
                \centering
                \includegraphics[width=\textwidth]{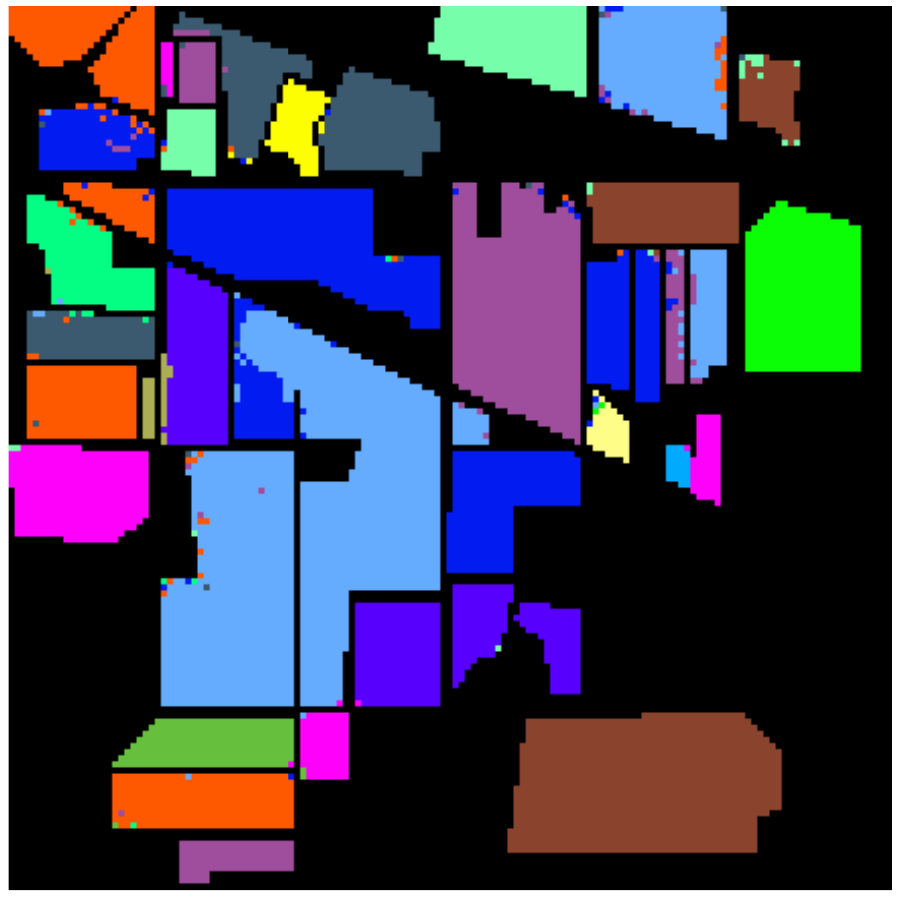}
                \caption{}
                \label{fig.indianCODL}
        \end{subfigure}
        \begin{subfigure}{0.125\textheight}
                \centering
                \includegraphics[width=\textwidth]{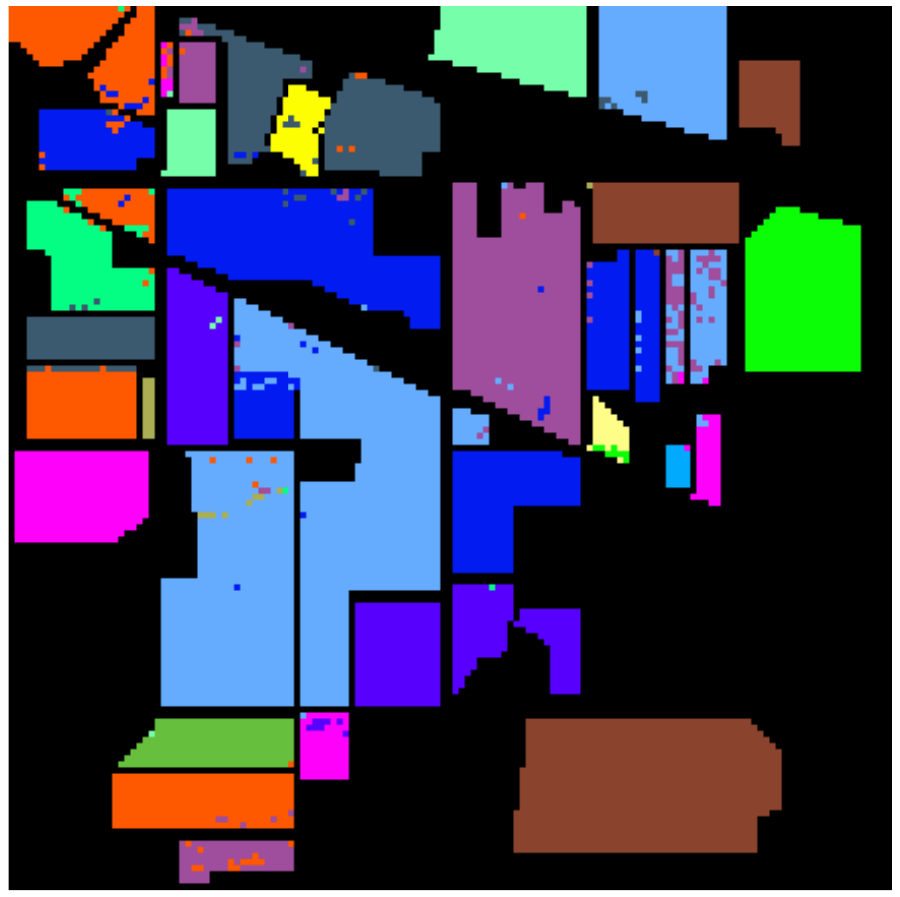}
                \caption{}
                \label{fig.indianLAJSR}
        \end{subfigure}
                \begin{subfigure}{0.125\textheight}
                \centering
                \includegraphics[width=\textwidth]{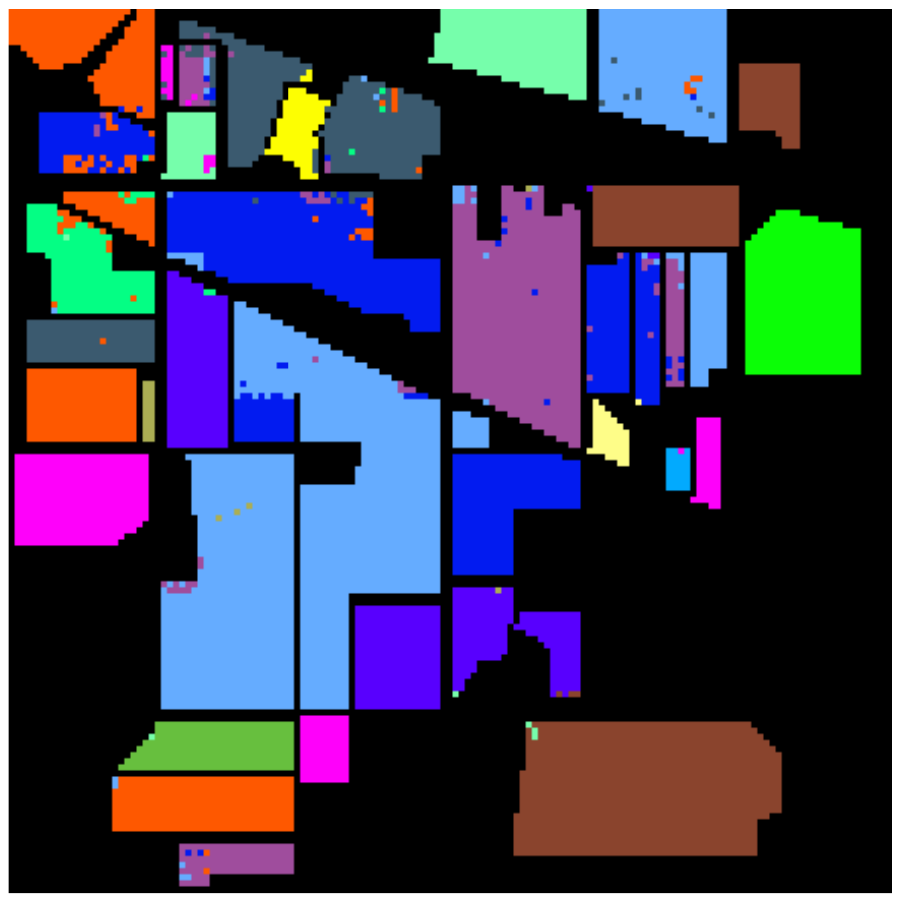}
                \caption{}
                \label{fig.indianSMSB}
        \end{subfigure}
        \caption{Classification results for Indian Pines image. (a) Ground truth. (b) Training data. (c) Test data. (d) SVM \cite{melgani04}. (e) SOMP \cite{chen11}. (f) SADL \cite{soltani15}. (g) LGIDL \cite{he16}. (h) CODL \cite{fu18}. (i) LAJSR \cite{peng2019}. (j) SMSB.}
        \label{fig.indianresult}
\end{figure}
\begin{table}
  \centering
\caption{Detailed classification results including class-specific accuracy (\%), overall accuracy (OA), average accuracy (AA), $\kappa$ coefficient and processing time (sec) for Indian Pines image with different classification approaches. }
\resizebox{0.96\textwidth}{0.2\textheight}{%
\label{table.Indian}
\begin{tabular}{|c|c|c c|c |c |c |c |c |c |c|}
  \hline
  Class & Name & Train & Test & SVM & SOMP & SADL & LGIDL & CODL &LAJSR & SMSB \\
  \hline \hline
  {\cellcolor [rgb]{0.9961,0.9922,0.5352}1} & Alfalfa & 6 & 40 & 82.50 \!$\pm 1.31$&57.50 \!$\pm 1.81$& \textbf{100} \!$\pm 0$& 85.00 \!$\pm 0.81$& 89.58 \!$\pm 0.78$& 80 \!$\pm 1.47$&	\textbf{100} \!$\pm 0$\\
  {\cellcolor [rgb]{0.0117,0.1094,0.9414}2} & Corn-no till & 153 & 1275 & 84.23 \!$\pm 0.81$& 88.00 \!$\pm 0.73$& 92.86 \!$\pm 0.47$& 95.37 \!$\pm 1.02$& 97.50 \!$\pm 0.45$& \textbf{97.57} \!$\pm 0.53$& 93.18 \!$\pm 0.44$\\
  {\cellcolor [rgb]{0.9961,0.3477,0.0039}3} & Corn-min till & 84 & 746 & 78.01 \!$\pm 0.43$& 81.09 \!$\pm 0.98$& 92.09 \!$\pm 0.23$& 94.24 \!$\pm 0.87$& \textbf{98.67} \!$\pm 0.53$& 94.23 \!$\pm 0.79$& 98.52 \!$\pm 0.13$\\
  {\cellcolor [rgb]{0.0195,0.9961,0.5195}4} & Corn & 28 & 245 & 76.55 \!$\pm 1.23$& 77.51 \!$\pm 0.22$& 95.69 \!$\pm 0.80$ & 88.00 \!$\pm 0.48$& 96.12 \!$\pm 0.31$& \textbf{96.17} \!$\pm 0.40$& 91.87 \!$\pm 0.43$\\
  {\cellcolor [rgb]{0.9961,0.0078,0.9805}5} & Grass/pasture-mowed & 48 & 435 & 94.71 \!$\pm 0.77$& 94.71 \!$\pm 0.12$& 91.26 \!$\pm 0.91$&	98.16 \!$\pm 0.55$& 98.44 \!$\pm 0.18$& 94.94 \!$\pm 0.79$& \textbf{99.31} \!$\pm 0.24$\\
  {\cellcolor [rgb]{0.3477,0.0039,0.9961}6} & Grass-trees & 64 & 666 & 93.54 \!$\pm 0.23$& 99.09 \!$\pm 0.04$& \textbf{99.85} \!$\pm 0.11$&	99.55 \!$\pm 0.04$& 97.22 \!$\pm 0.34$& 99.52 \!$\pm 0.09$& 98.95 \!$\pm 0.77$\\
  {\cellcolor [rgb]{0.0117,0.6680,0.9961}7} & Grass/pasture & 4 & 24 & 87.50 \!$\pm 0.41$& 95.83 \!$\pm 0.51$& \textbf{100} \!$\pm 0$& 95.81 \!$\pm 0.67$& 95.45 \!$\pm 1.21$& 95.83 \!$\pm 0.31$& 95.79 \!$\pm 0.11$\\
  {\cellcolor [rgb]{0.0469,0.9961,0.0273}8} & Hay-windrowed & 48 & 430 & 98.83 \!$\pm 0.76$& 99.06 \!$\pm 0.05$& \textbf{100} \!$\pm 0$& \textbf{100} \!$\pm 0$& \textbf{100} \!$\pm 0$& \textbf{100} \!$\pm 0$& \textbf{100} \!$\pm 0$\\
  {\cellcolor [rgb]{0.6719,0.6836,0.3281}9} & Oats & 4 & 16 &87.50 \!$\pm 0.57$& 75.00 \!$\pm 1.41$& \textbf{100} \!$\pm 0$& 93.75 \!$\pm 0.75$& \textbf{100} \!$\pm 0$& \textbf{100} \!$\pm 0$&\textbf{100} \!$\pm 0$\\
  {\cellcolor [rgb]{0.6250,0.3047,0.6172}10}& Soybean-no till & 96 & 876 & 78.31 \!$\pm 0.88$& 85.04 \!$\pm 0.47$& 96.57 \!$\pm 0.93$&93.61 \!$\pm 0.91$& \textbf{97.71} \!$\pm 0.22$& 93.83 \!$\pm 0.47$& 93.04 \!$\pm 0.08$\\
  {\cellcolor [rgb]{0.3979,0.6778,1}11}& Soybean-min till & 170 & 2285 & 79.21 \!$\pm 1.02$& 90.67 \!$\pm 0.61$& 95.75 \!$\pm 0.26$& 95.45 \!$\pm 0.52$& 95.26 \!$\pm 0.09$&	96.50 \!$\pm 0.36$& \textbf{98.34} \!$\pm 0.41$\\
  {\cellcolor [rgb]{0.2344,0.3555,0.4375}12}& Soybean-clean till & 63 & 530 & 81.70 \!$\pm 0.89$& 86.41 \!$\pm 0.45$& 90.18 \!$\pm 0.64$& 92.83 \!$\pm 0.23$& 95.83 \!$\pm 0.93$& \textbf{96.98} \!$\pm 0.53$& 96.23 \!$\pm 0.61$\\
  {\cellcolor [rgb]{0.4063,0.7500,0.2461}13}& Wheat & 18 & 187 & 95.72 \!$\pm 0.75$& 100 \!$\pm 0$& 98.39 \!$\pm 0.27$& \textbf{100} \!$\pm 0$& 99.48 \!$\pm 0.14$& 98.93 \!$\pm 0.82$& 99.46 \!$\pm 0.04$\\
  {\cellcolor [rgb]{0.5430,0.2695,0.1797}14}& Woods & 100 & 1165 &90.98 \!$\pm 0.24$&96.48 \!$\pm 0.23$& 98.97 \!$\pm 0.84$& 98.28 \!$\pm 0.53$& 98.74 \!$\pm 0.31$& \textbf{99.91} \!$\pm 0.03$& 99.66 \!$\pm 0.08$\\
  {\cellcolor [rgb]{0.4649,0.9961,0.6719}15}& Bldg-grass-trees-drives & 48 & 338 & 77.81 \!$\pm 0.89$& 92.60 \!$\pm 0.72$& 92.30 \!$\pm 0.93$&	95.56 \!$\pm 0.18$& 99.40 \!$\pm 0.71$& \textbf{100} \!$\pm 0$& 98.52 \!$\pm 0.23$\\
  {\cellcolor [rgb]{0.9922,0.9961,0.0117}16}& Stone-steel-towers & 8 & 85 &97.65 \!$\pm 0.46$& \textbf{100} \!$\pm 0$& 92.94 \!$\pm 0.49$& \textbf{100} \!$\pm 0$& \textbf{100} \!$\pm 0$& 91.76 \!$\pm 0.52$&94.12 \!$\pm 0.31$\\
  \hline \hline
  OA& - & - & - & 84.43 \!$\pm 1.20$& 90.56 \!$\pm 0.85$& 95.44 \!$\pm 0.71$& 95.94 \!$\pm 0.51$& \textbf{97.33} \!$\pm 0.66$& 97.03 \!$\pm 0.68$& 97.21 \!$\pm 0.41$\\
  \hline
  AA& - & - & - & 86.55 \!$\pm 0.61$& 88.69 \!$\pm 0.71$& 96.05 \!$\pm 0.62$& 95.40 \!$\pm 0.44$& \textbf{97.46} \!$\pm 0.53$& 96.01 \!$\pm 0.62$& 97.31 \!$\pm 0.35$\\
  \hline
  $\kappa$& - & - & - & 82.28 \!$\pm 0.99$& 89.20 \!$\pm 0.81$& 94.79 \!$\pm 0.66$& 95.36 \!$\pm 0.61$& \textbf{96.95} \!$\pm 0.59$& 96.61 \!$\pm 0.56$& 96.81 \!$\pm 0.42$\\
  \hline \hline
  Time (sec) & - & - & - & 5 \!$\pm 1$& 95 \!$\pm 5$& 61 \!$\pm 3$& 103 \!$\pm 6$& 58 \!$\pm 3$&158 \!$\pm 4$& \textbf{12} \!$\pm 1$\\
  \hline
\end{tabular}
}
\end{table}
\textit{Indian Pines Image:} \hyperref[fig.indian]{Figure \ref*{fig.indian}} presents the ground truth image of the Indian Pines dataset which contains 16 different classes, and the colors are specified in \hyperref[table.Indian]{Table \ref*{table.Indian}}. The exact numbers of train and test data for each class are presented in \hyperref[table.Indian]{Table \ref*{table.Indian}} and visualized in \hyperref[fig.indianresult]{Figure \ref*{fig.indianresult}}. The detailed classification accuracies and processing time of each method are presented in \hyperref[table.Indian]{Table \ref*{table.Indian}}, and the visual classification results are depicted in \hyperref[fig.indianresult]{Figure \ref*{fig.indianresult}}. The reported values in \hyperref[table.Indian]{Table \ref*{table.Indian}} are the mean and standard deviation values obtained from ten runs.

Similar to the Pavia University results, for Indian Pines image, all of the reviewed methods outperform SVM due to their additional feature extraction step. Training a dictionary notably improves the classification performance in SADL, LGIDL, CODL and SMSB approaches. However, for Indian Pines image, the LAJSR method achieves better performance (compared to SADL and LGIDL) without training a dictionary. This can be explained via the use of overlapping spatial patches in LAJSR, which can successfully exploit the spatial correlations of the homogenous regions of Indian Pines image. For Indian Pines image, the CODL method outperforms the SMSB method with 0.47\% higher accuracy. However, achieving this accuracy takes more than four times more processing time.

Analyzing the detailed classification results reveals that the proposed SMSB method has failed to achieve the best classification results at some certain classes, especially the \emph{Corn} (91.87\%) and \emph{Stone-steal-towers} (94.12\%). It can be observed from \hyperref[fig.indianSMSB]{Figure \ref*{fig.indianSMSB}} that the majority of classification errors of the SMSB method occur at the edges and they are mostly misclassified as the \emph{Corn-min-till} class. This can be explained as the edge problem of the SMSB method and it is the result of non-overlapping spatial patches that are used in this method. The LAJSR and the CODL methods appear to be robust to the edge problem at \emph{Corn} class because of the utilization of the overlapping patches and extra regularization penalty, respectively. Similar edge errors occur at the \emph{Stone-steel-towers} class for SMSM method, which are mostly misclassified as the \emph{Soybean-clean till} class.
\begin{figure}
        \centering

        \begin{subfigure}{0.125\textheight}
                \centering
                \includegraphics[width=\textwidth]{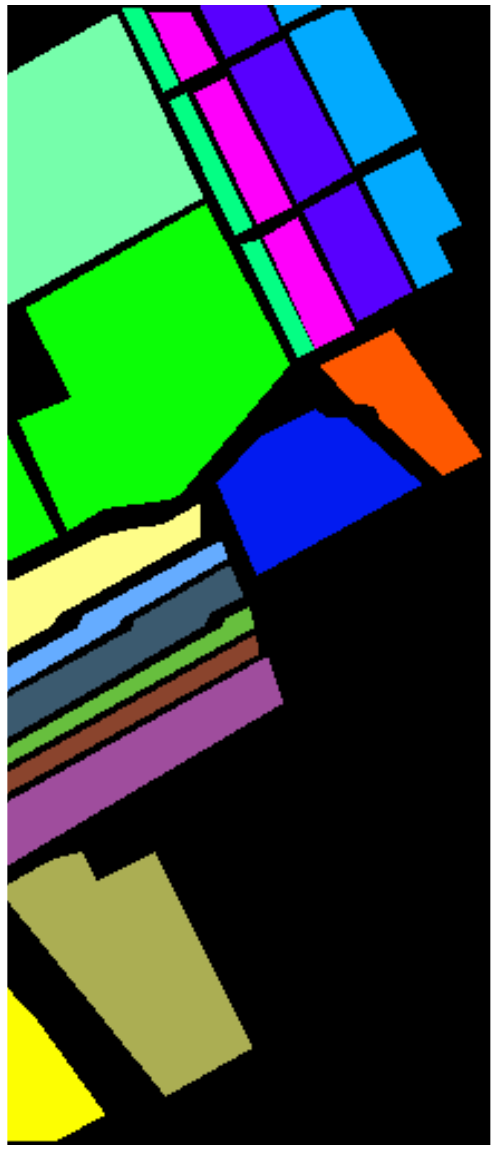}
                \caption{}
                \label{fig.salinas}
        \end{subfigure}
        \begin{subfigure}{0.125\textheight}
                \centering
                \includegraphics[width=\textwidth]{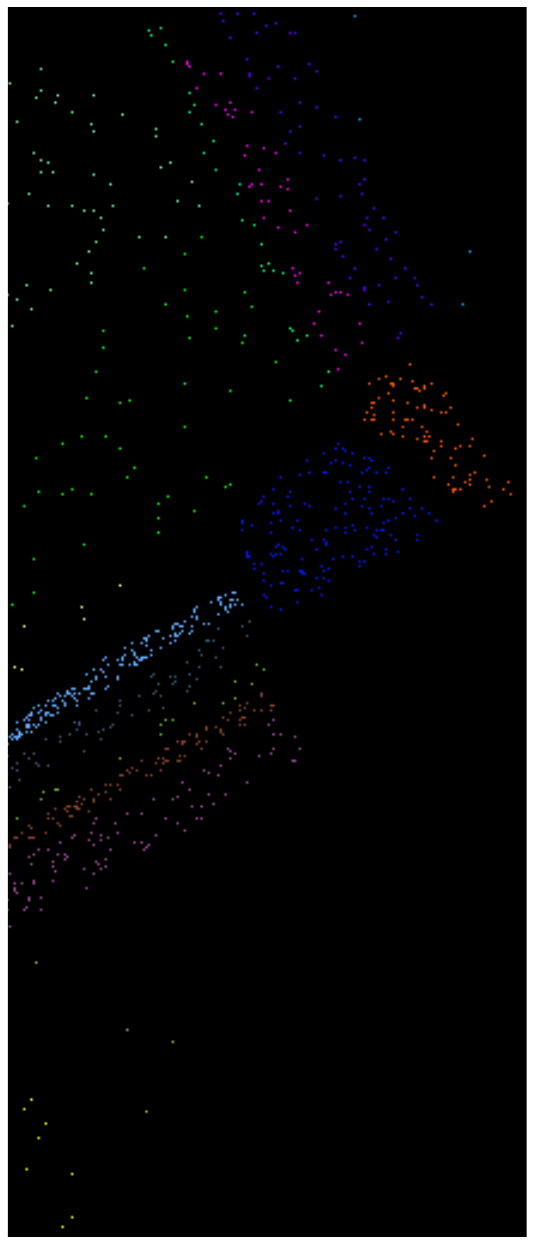}
                \caption{}
                \label{fig.salinastrain}
        \end{subfigure}
        \begin{subfigure}{0.125\textheight}
                \centering
                \includegraphics[width=\textwidth]{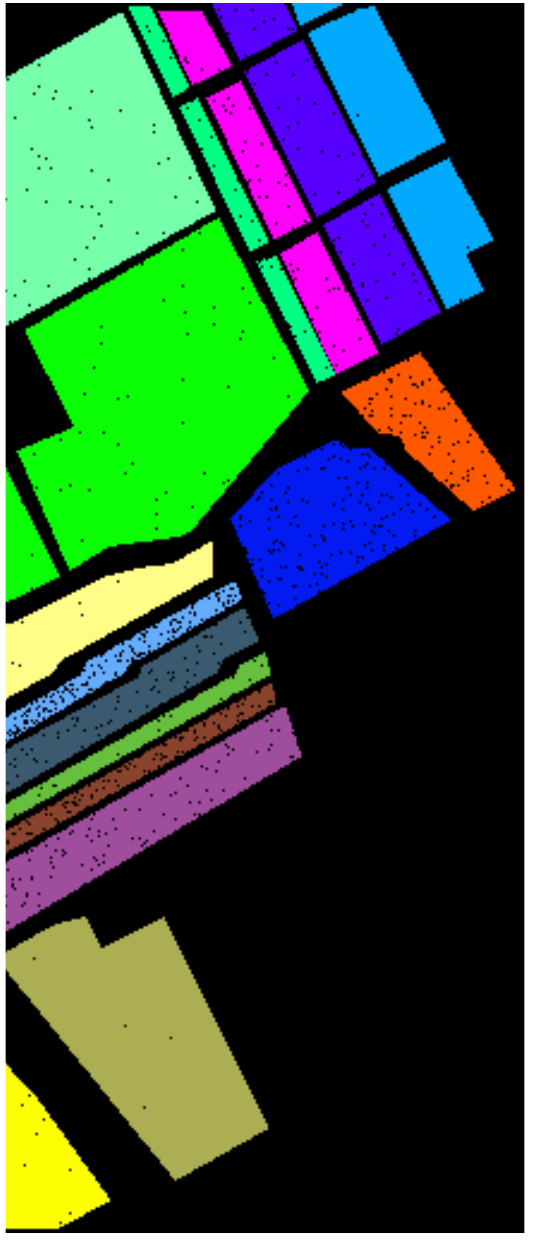}
                \caption{}
                \label{fig.salinastest}
        \end{subfigure}
        \begin{subfigure}{0.125\textheight}
                \centering
                \includegraphics[width=\textwidth]{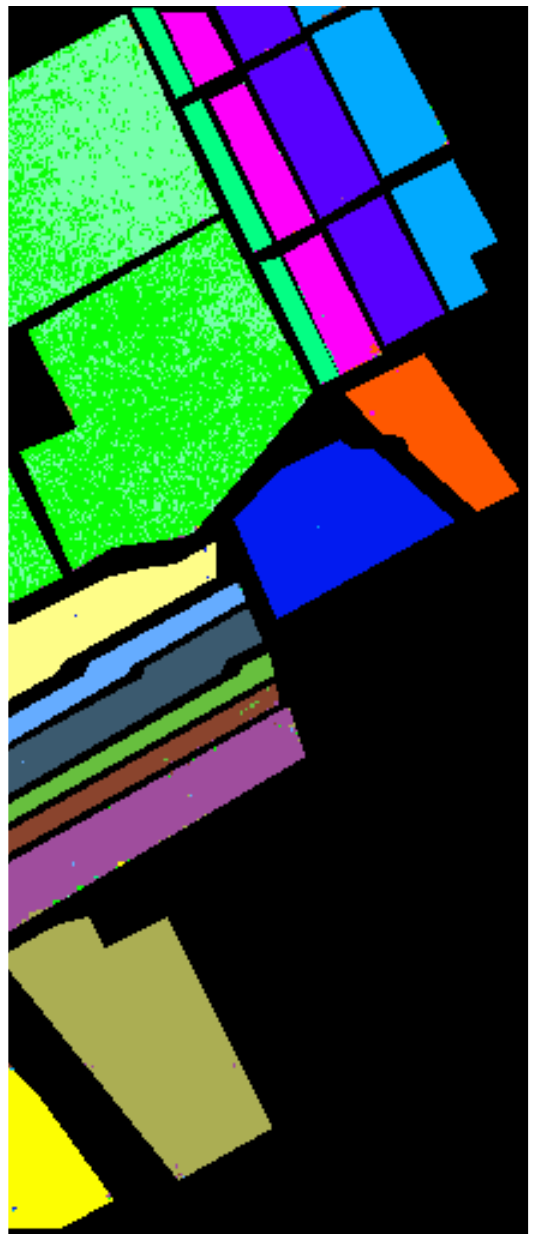}
                \caption{}
                \label{fig.salinasSVM}
        \end{subfigure}
        \begin{subfigure}{0.125\textheight}
                \centering
                \includegraphics[width=\textwidth]{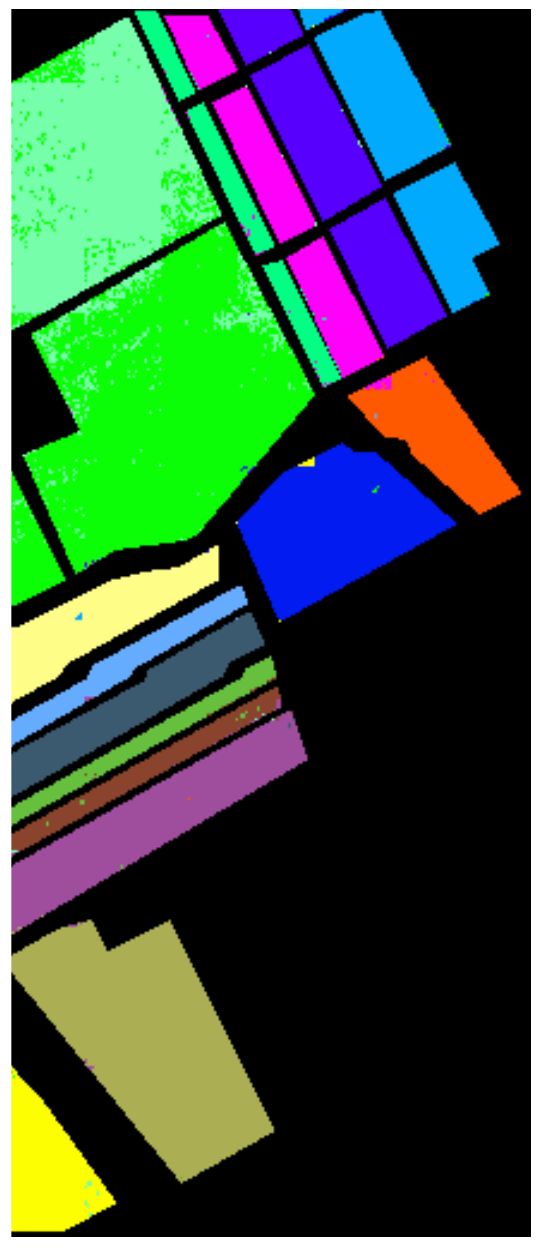}
                \caption{}
                \label{fig.salinasSOMP}
        \end{subfigure}
        \begin{subfigure}{0.125\textheight}
                \centering
                \includegraphics[width=\textwidth]{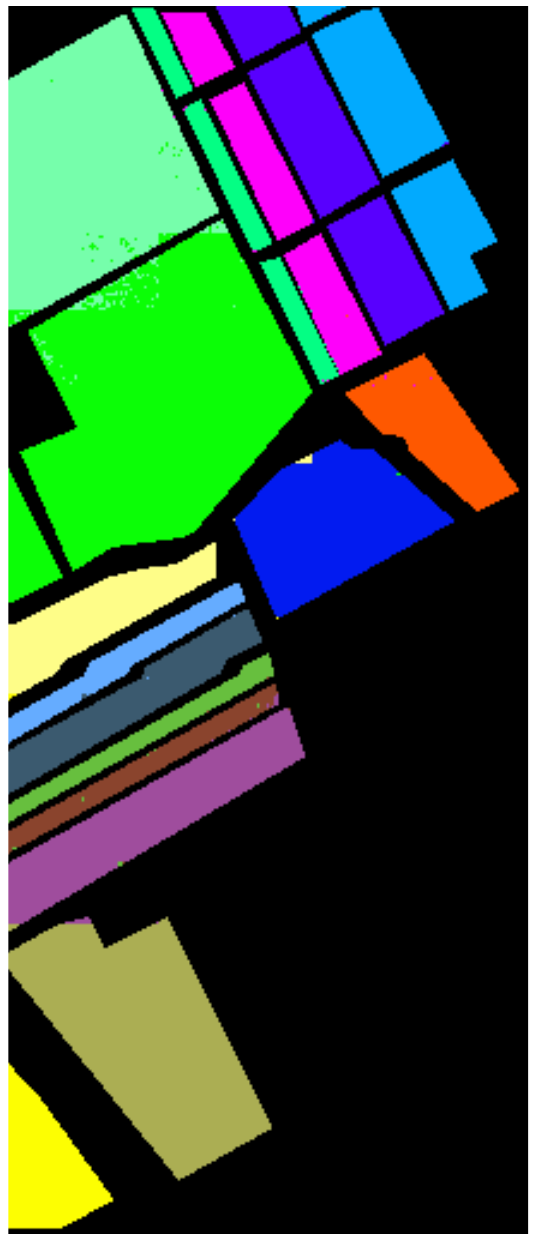}
                \caption{}
                \label{fig.salinasSADL}
        \end{subfigure}
        \begin{subfigure}{0.125\textheight}
                \centering
                \includegraphics[width=\textwidth]{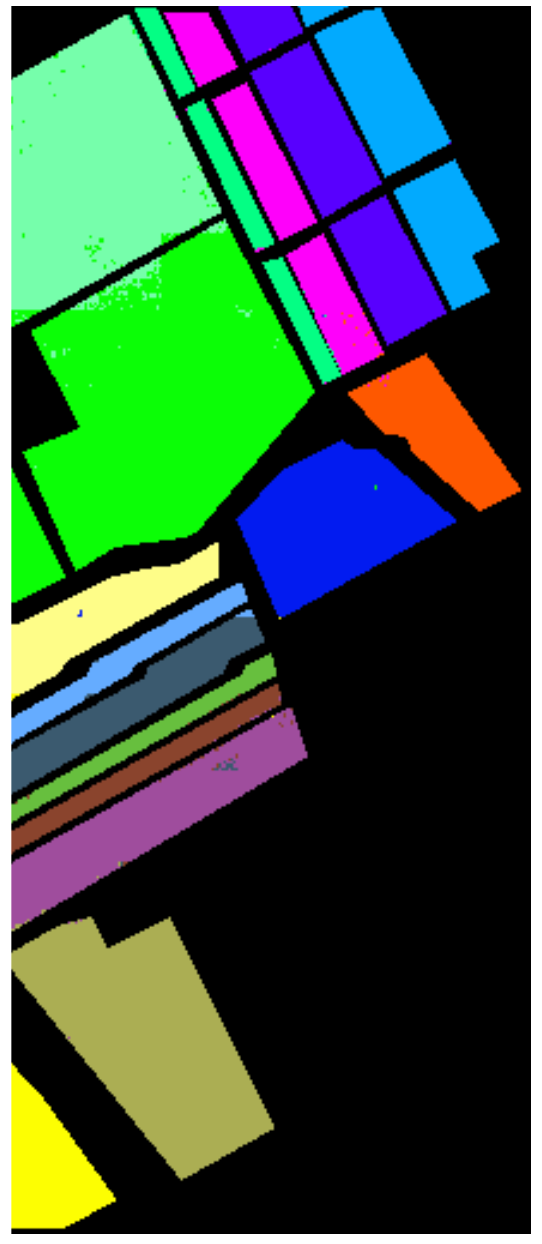}
                \caption{}
                \label{fig.salinasLGIDL}
        \end{subfigure}
        \begin{subfigure}{0.125\textheight}
                \centering
                \includegraphics[width=\textwidth]{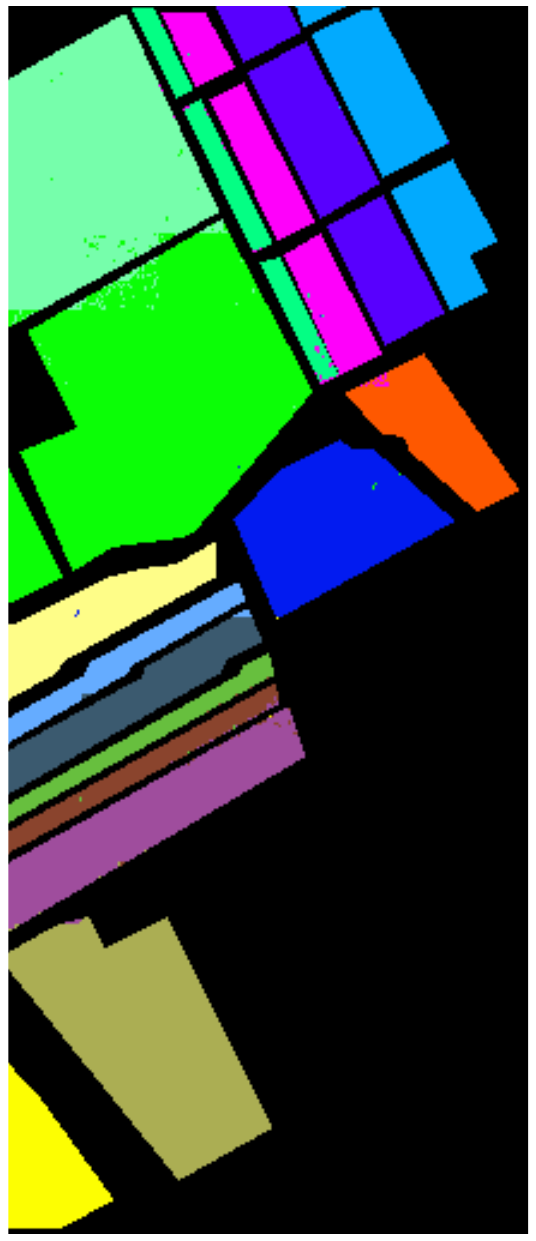}
                \caption{}
                \label{fig.salinasCODL}
        \end{subfigure}
        \begin{subfigure}{0.125\textheight}
                \centering
                \includegraphics[width=\textwidth]{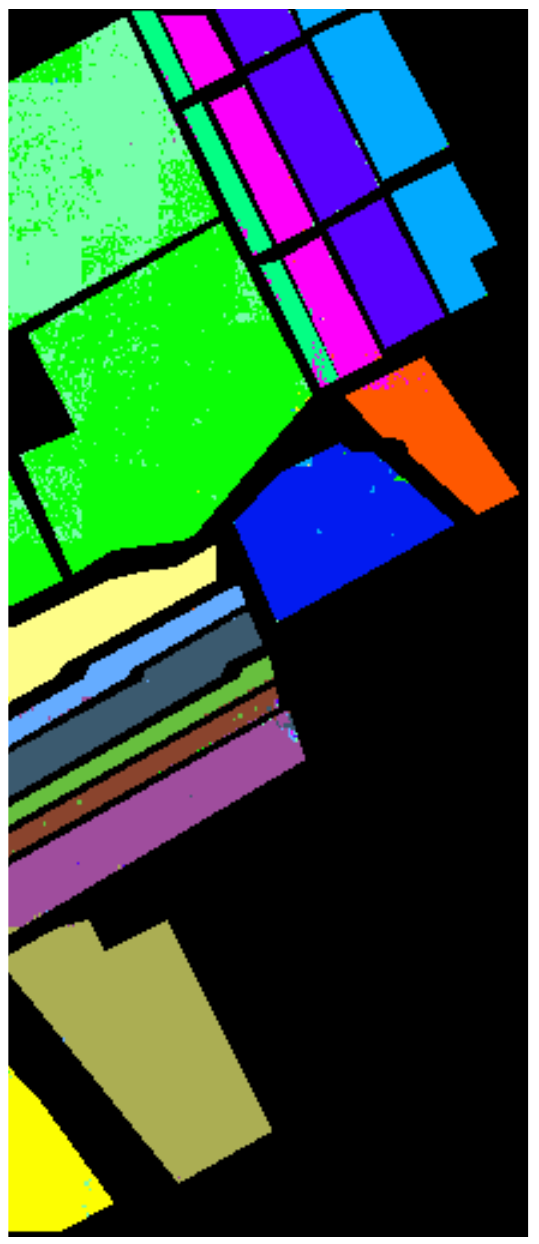}
                \caption{}
                \label{fig.salinasLAJSR}
        \end{subfigure}
                \begin{subfigure}{0.125\textheight}
                \centering
                \includegraphics[width=\textwidth]{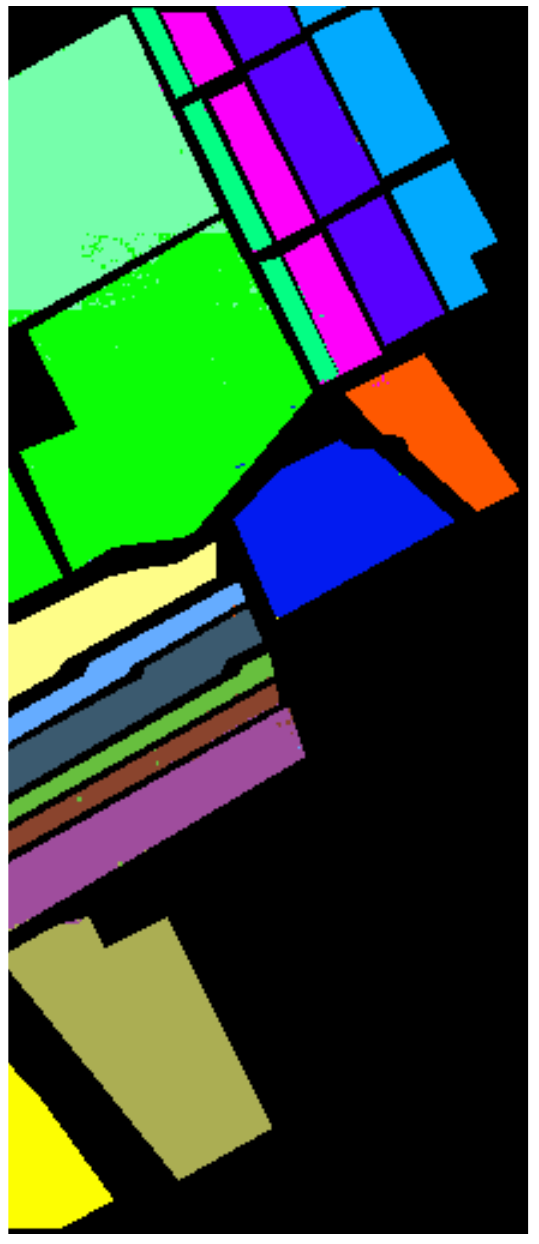}
                \caption{}
                \label{fig.salinasSMSB}
        \end{subfigure}
        \caption{Classification results for Salinas image. (a) Ground truth. (b) Training data. (c) Test data. (d) SVM \cite{melgani04}. (e) SOMP \cite{chen11}. (f) SADL \cite{soltani15}. (g) LGIDL \cite{he16}. (h) CODL \cite{fu18}. (i) LAJSR \cite{peng2019}. (j) SMSB.}
        \label{fig.salinasresult}
\end{figure}

\begin{table}
  \centering
\caption{Detailed classification results including class-specific accuracy (\%), overall accuracy (OA), average accuracy (AA), $\kappa$ coefficient and processing time (sec) for Salinas image with different classification approaches. }
\resizebox{0.98\textwidth}{0.25\textheight}{%
\label{table.Salinas}
\begin{tabular}{|c|c|c c|c |c| c| c| c| c| c|}
  \hline
  Class & Name & Train & Test & SVM & SOMP & SADL & LGIDL & CODL &LAJSR & SMSB \\
  \hline \hline
  {\cellcolor [rgb]{0.9961,0.9922,0.5352}1} & Brocoli-green-weeds-1 & 201 & 1808 & 99.51 \!$\pm 0.43$ &	99.68 \!$\pm 0.22$& 99.57 \!$\pm 0.3$& 99.69 \!$\pm 0.36$& 99.55 \!$\pm 0.27$& 99.57 \!$\pm 0.33$&	\textbf{99.78 \!$\pm 0.2$} \\
  {\cellcolor [rgb]{0.0117,0.1094,0.9414}2} & Brocoli-green-weeds-2 & 372 & 3354 & 99.84 \!$\pm 0.20$ & 99.38 \!$\pm 0.31$ & 99.84 \!$\pm 0.19$& 99.73 \!$\pm 0.23$& 99.83 \!$\pm 0.18$& 99.59 \!$\pm 0.26$& \textbf{99.97} \!$\pm 0.05$ \\
  {\cellcolor [rgb]{0.9961,0.3477,0.0039}3} & Fallow & 197 & 1779 & 96.92 \!$\pm 7.61$&	97.25 \!$\pm 0.84$& 99.49 \!$\pm 0.8$& 98.75 \!$\pm 1.0$& 99.73 \!$\pm 0.31$& 97.96 \!$\pm 1.11$& \textbf{99.94} \!$\pm 0.11$ \\
  {\cellcolor [rgb]{0.0195,0.9961,0.5195}4} & Fallow-rough-plow & 139 & 1255 & 98.45 \!$\pm 3.06$ & 96.46 \!$\pm 1.62$ & 99.04 \!$\pm 0.57$ & 98.45 \!$\pm 1.0$& 98.88 \!$\pm 0.32$& 98.41 \!$\pm 1.08$& \textbf{99.28} \!$\pm 0.38$\\
  {\cellcolor [rgb]{0.9961,0.0078,0.9805}5} & Fallow-smooth & 268 & 2410 & 98.21 \!$\pm 0.32$& 98.72 \!$\pm 0.3$& 99.12 \!$\pm 0.36$&	98.89 \!$\pm 0.43$& 99.24 \!$\pm 0.22$& 98.94 \!$\pm 0.47$& \textbf{99.54} \!$\pm 0.23$\\
  {\cellcolor [rgb]{0.3477,0.0039,0.9961}6} & Stubble & 396 & 3563 & 99.73 \!$\pm 0.14$& 99.83 \!$\pm 0.11$& 99.93 \!$\pm 0.08$& 99.89 \!$\pm 0.13$& \textbf{99.98} \!$\pm 0.02$& 99.86 \!$\pm 0.13$& 99.97 \!$\pm 0.02$\\
  {\cellcolor [rgb]{0.0117,0.6680,0.9961}7} & Celery & 358 & 3221 & 99.67 \!$\pm 0.18$& 99.60 \!$\pm 0.18$& \textbf{99.92} \!$\pm 0.08$& 99.73 \!$\pm 0.20$& 99.91 \!$\pm 0.08$& 99.63 \!$\pm 0.22$& 99.88 \!$\pm 0.10$\\
  {\cellcolor [rgb]{0.0469,0.9961,0.0273}8} & Grapes-untrained & 1127 & 10144 & 79.78 \!$\pm 4.43$&	92.32 \!$\pm 0.38$& 98.42 \!$\pm 0.46$& 96.59 \!$\pm 1.20$& 98.69 \!$\pm 0.12$& 95.52 \!$\pm 2.72$& \textbf{98.87} \!$\pm 0.13$\\
  {\cellcolor [rgb]{0.6719,0.6836,0.3281}9} & Soil-vinyard-develop & 620 & 5583 &99.61 \!$\pm 0.66$& 99.80 \!$\pm 0.18$ & 99.77 \!$\pm 0.13$& 99.74 \!$\pm 0.19$& 99.85 \!$\pm 0.13$& 99.77 \!$\pm 0.20$&\textbf{99.91} \!$\pm 0.05$ \\
  {\cellcolor [rgb]{0.6250,0.3047,0.6172}10}& Corn-senesced-green-weeds & 328 & 2950 & 97.43 \!$\pm 0.74$& 97.98 \!$\pm 0.79$& 98.80 \!$\pm 0.63$& 98.57 \!$\pm 0.49$& 98.43 \!$\pm 0.67$& 98.42 \!$\pm 0.66$& \textbf{98.85} \!$\pm 0.54$\\
  {\cellcolor [rgb]{0.3979,0.6778,1}11}& Lettuce-romaine-4wk & 107 & 961 & 99.23 \!$\pm 0.42$ & 99.04  \!$\pm 0.70$& 99.40 \!$\pm 0.32$& 99.50 \!$\pm 0.40$& 99.30 \!$\pm 0.51$&99.47 \!$\pm 0.38$& \textbf{99.79} \!$\pm 0.33$\\
  {\cellcolor [rgb]{0.2344,0.3555,0.4375}12}& Lettuce-romaine-5wk & 192 & 1735 & 99.85 \!$\pm 0.21$& 99.58 \!$\pm 0.42$&	99.80 \!$\pm 0.32$& 99.56 \!$\pm 0.47$& 99.79 \!$\pm 0.35$& 99.63 \!$\pm 0.43$&\textbf{99.94} \!$\pm 0.02$ \\
  {\cellcolor [rgb]{0.4063,0.7500,0.2461}13}& Lettuce-romaine-6wk & 91 & 825 & 98.68 \!$\pm 1.57$&98.50 \!$\pm 1.09$& 98.66 \!$\pm 1.09$&98.25 \!$\pm 1.49$& \textbf{99.25} \!$\pm 0.40$&98.20 \!$\pm 1.32$& 99.03 \!$\pm 0.43$ \\
  {\cellcolor [rgb]{0.5430,0.2695,0.1797}14}& Lettuce-romaine-7wk & 107 & 963 &97.33 \!$\pm 1.45$&96.28 \!$\pm 0.93$ &97.76 \!$\pm 0.88$& 97.71 \!$\pm 0.90$& \textbf{99.97} \!$\pm 0.66$& 97.59 \!$\pm 1.31$& 98.86 \!$\pm 0.57$ \\
  {\cellcolor [rgb]{0.4649,0.9961,0.6719}15}& Vinyard-untrained & 326 & 6942 &68.69 \!$\pm 7.5$& 83.49 \!$\pm 3.8$& 97.85 \!$\pm 0.28$&	93.23 \!$\pm 4.20$& \textbf{97.93} \!$\pm 0.33$& 91.04 \!$\pm 6.02$& 97.63 \!$\pm 0.30$ \\
  {\cellcolor [rgb]{0.9922,0.9961,0.0117}16}& Vinyard-vertical-trellis & 180 & 1627 &98.62 \!$\pm 1.32$& 98.81 \!$\pm 0.86$& 99.58 \!$\pm 0.54$& 98.73 \!$\pm 0.74$& 99.72 \!$\pm 0.61$& 98.60 \!$\pm 0.86$& \textbf{99.92} \!$\pm 0.03$ \\
  \hline \hline
  OA& - & - & - & 90.93 \!$\pm 1.29$& 95.52 \!$\pm 0.62$& 99.06 \!$\pm 0.11$& 97.93 \!$\pm 0.91$& 99.14 \!$\pm 0.05$& 97.35 \!$\pm 1.47$& \textbf{99.26} \!$\pm 0.11$ \\
  \hline
  AA& - & - & - & 95.72 \!$\pm 0.55$& 97.29 \!$\pm 0.47$& 99.18 \!$\pm 0.08$& 98.56 \!$\pm 0.51$&99.25 \!$\pm 0.04$& 98.26 \!$\pm 0.78$& \textbf{99.45} \!$\pm 0.07$\\
  \hline
  $\kappa$& - & - & - & 89.90 \!$\pm 1.44$& 95.01 \!$\pm 0.69$&98.95 \!$\pm 0.12$& 97.69 \!$\pm 1.02$ & 99.04 \!$\pm 0.05$& 97.05 \!$\pm 1.64$& \textbf{99.17} \!$\pm 0.05$\\
  \hline \hline
  Time (sec) & - & - & - & 45 \!$\pm 1$& 126 \!$\pm 4$& 292 \!$\pm 7$& 406 \!$\pm 8$& 178 \!$\pm 5$& 660 \!$\pm 10$& \textbf{51} \!$\pm 3$\\
  \hline
\end{tabular}
}
\end{table}
\textit{Salinas Image:} To further analyze the performance of the proposed method, it is applied on the Salinas dataset. \hyperref[fig.salinas]{Figure \ref*{fig.salinas}} presents the ground truth image of the Salinas dataset which contains 16 different classes, and the colors are specified in \hyperref[table.Salinas]{Table \ref*{table.Salinas}}. The exact numbers of train and test data for each class are presented in \hyperref[table.Salinas]{Table \ref*{table.Salinas}} and visualized in \hyperref[fig.salinasresult]{Figure \ref*{fig.salinasresult}}. The detailed classification accuracies and processing time of each method are presented in \hyperref[table.Salinas]{Table \ref*{table.Salinas}}, and the visual classification results are depicted in \hyperref[fig.salinasresult]{Figure \ref*{fig.salinasresult}}. The reported values in \hyperref[table.Salinas]{Table \ref*{table.Salinas}} are the mean and standard deviation values obtained from ten runs.

The Salinas dataset is a challenging dataset in terms of its processing time, since it contain a large number of pixels. Hence, it is a good example to verify the effectiveness of the proposed method in reducing the processing time, while reaching a good performance. The results obtained for the Salinas dataset are very similar to the results obtained for the other two datasets. It is observable in \hyperref[table.Salinas]{Table \ref*{table.Salinas}} that except for SVM and SOMP, the rest of the examined methods achieve an overall accuracy of more than 95\% and the proposed SMSM method outperform the rest of the methods with the accuracy of 99.26\%. The important fact here is that the SMSB method achieves this accuracy in only 51 seconds, which is very low compared to the rest of the methods. This verifies the effectiveness of the proposed spectral blocks and the selective spectral blocks in improving the performance of the HSI classification.
\begin{figure}
        \centering

        \begin{subfigure}[b]{0.16\textheight}
                \centering
                \includegraphics[width=\textwidth]{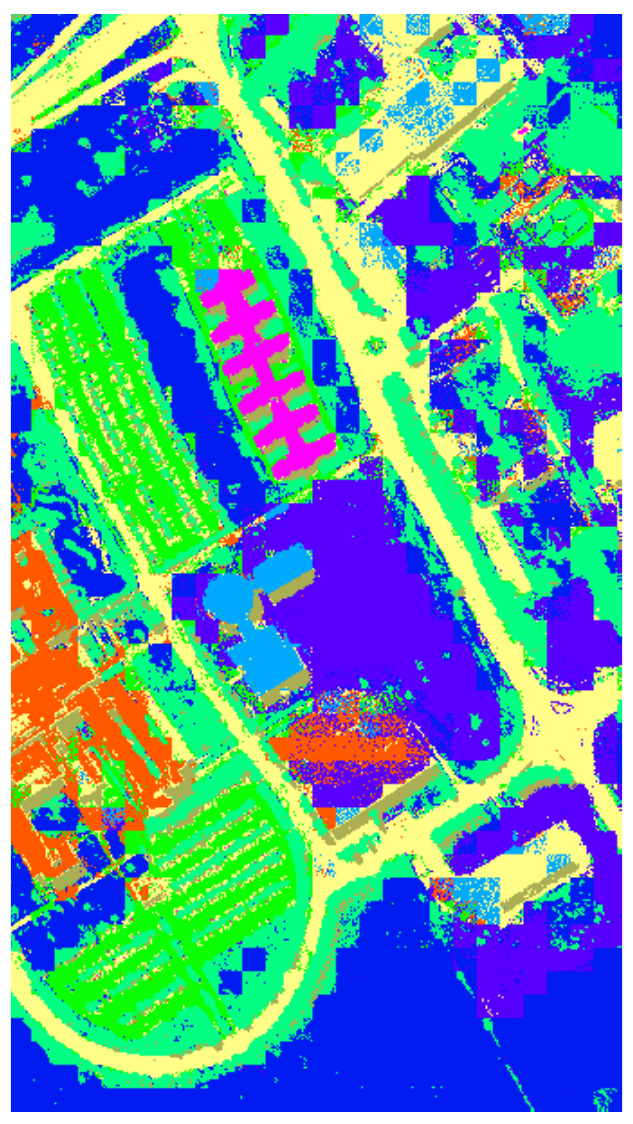}
                \caption{}
                \label{fig.indianT}
        \end{subfigure}
        \begin{subfigure}[b]{0.16\textheight}
                \centering
                \includegraphics[width=\textwidth]{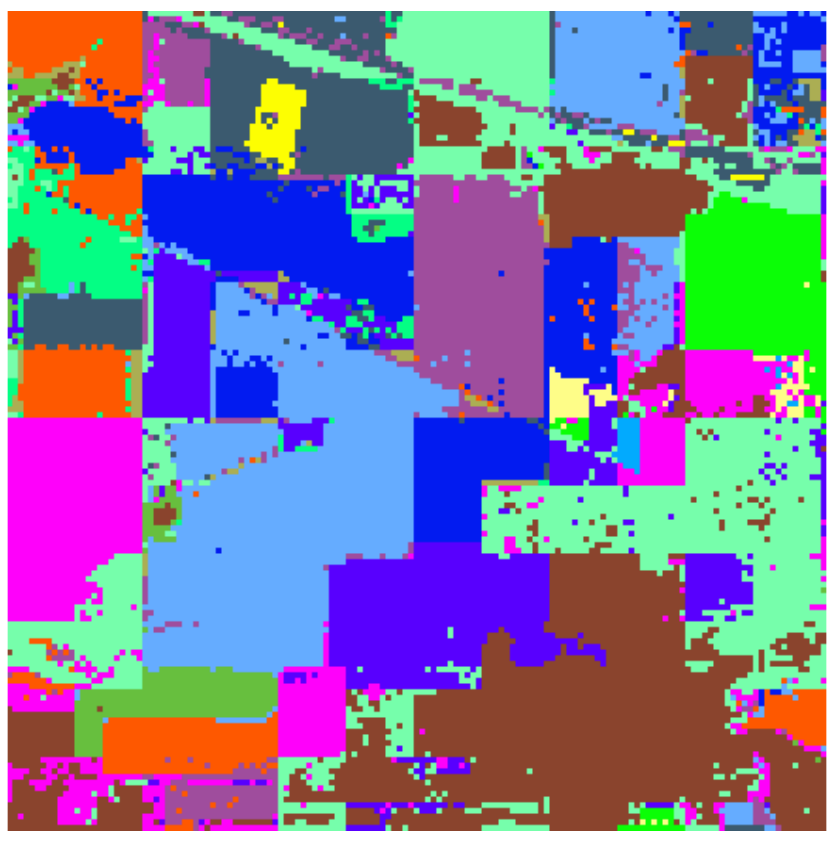}
                \caption{}
                \label{fig.paviaT}
        \end{subfigure}
        \begin{subfigure}[b]{0.16\textheight}
                \centering
                \includegraphics[width=\textwidth]{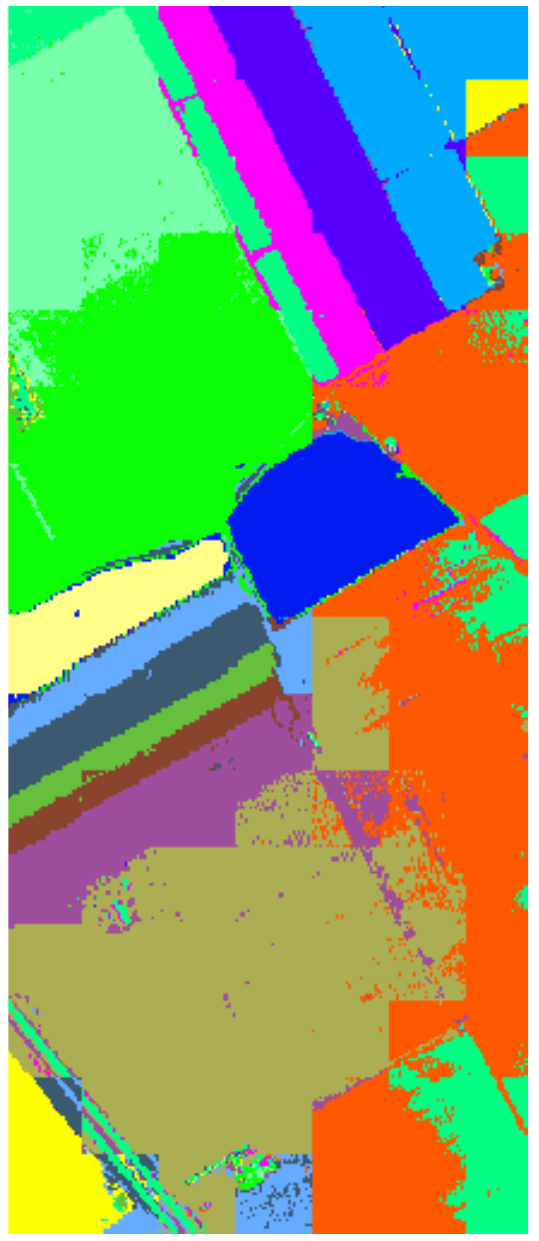}
                \caption{}
                \label{fig.salinasT}
        \end{subfigure}
        \caption{Complete classification map of (a) Pavia University image, (b) Indian Pines image and (c) Salinas image obtained by the proposed SMSB method.}
        \label{fig.cmaps}
\end{figure}

As it can be observed from \hyperref[table.Pavia]{Table \ref*{table.Pavia}}, \hyperref[table.Indian]{Table \ref*{table.Indian}} and \hyperref[table.Salinas]{Table \ref*{table.Salinas}}, many pixels of the aforementioned HSIs do not have ground-truth data. Therefore, they are represented in black as background. To make a better qualitative analysis of the proposed method, the complete classification maps of the Pavia University, Indian Pines and Salinas images are presented in \hyperref[fig.cmaps]{Figure \ref*{fig.cmaps}}.

\subsection{Processing Time}
While the computational complexity of different sparse modeling algorithms depends on their specific implementation details, the two important factors are the number of rows and columns of the dictionary, i.e. the dimensions of the problem. For instance, for an $m\times n$ dictionary, the basic OMP algorithm with $K$ iterations requires $2Kmn+3K^2m$ floating point operations (flops)\cite{wang2012}. That is, the computational complexity is of order $\mathrm{O}(mn)$.

Consider the Indian Pines image with 200 spectral bands ($m=200$). Instead of handling a problem with $m=200$ and $n>200$, the proposed SMSB method breaks it into 8 small-size problems with $m=20$ and $n=28$ (refer to \hyperref[table.tune]{Table \ref*{table.tune}}). Therefore, using the spectral blocks, the proposed SMSB method notably reduces the computational complexity of the sparse representation step. Note that compared to the other sparse modeling approaches (SADL, LGIDL and CODL), The SMSB trains a much smaller dictionary resulting in a faster dictionary learning step.

\begin{figure}
 \centering
 \includegraphics[width=0.7\textwidth]{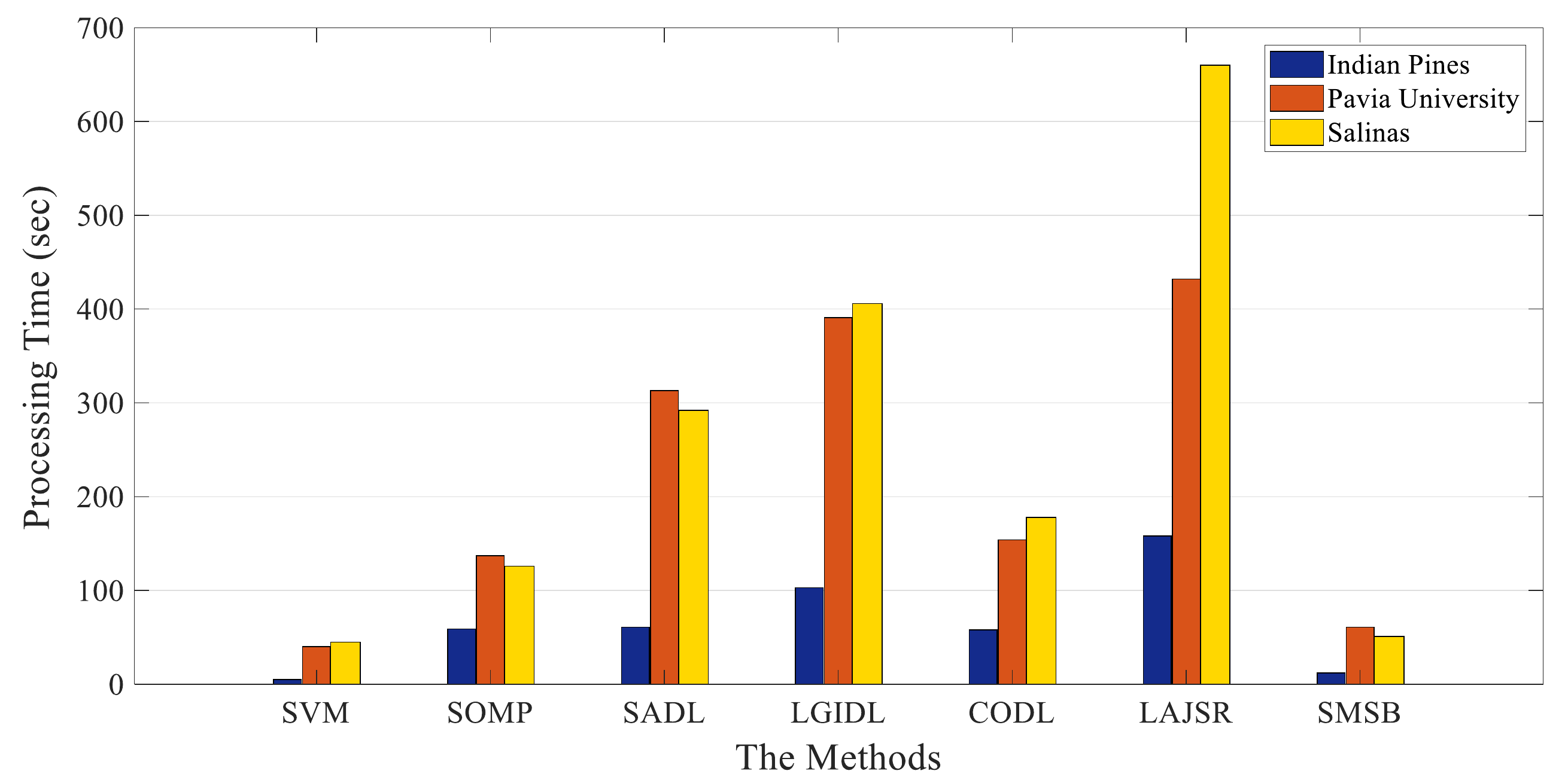}
 \caption{Processing time of the SVM, SVMCK, SOMP, SADL, LGIDL, CODL and SMSB methods for Indian Pines, Pavia University and Salinas images.}\label{fig.time}
\end{figure}

\hyperref[fig.time]{Figure \ref*{fig.time}} shows the processing time of the SVM, SOMP, SADL, LGIDL, CODL, LAJSR and SMSB methods for Indian Pines, Pavia University and Salinas images. A laptop computer with Intel Corei7(2.1 GHz) processor and 4 GB RAM was used for the implementation of the methods. For Pavia University Image, as it can be seen in \hyperref[fig.time]{Figure \ref*{fig.time}} and \hyperref[table.Pavia]{Table \ref*{table.Pavia}}, the proposed SMSB method outperforms the other methods in terms of both classification accuracy and processing time. For Indian Pines Image, according to \hyperref[fig.time]{Figure \ref*{fig.time}} and \hyperref[table.Indian]{Table \ref*{table.Indian}}, the proposed method outperforms SVM, SOMP, SADL LGIDL and LAJSR methods in terms of both classification accuracy and processing time. Although, the CODL method outperforms the proposed method with 0.47\% higher accuracy, the processing time (58 sec) is more than four times the processing time of the proposed SMSB method (12 sec). Salinas image contains a large number of pixels. therefore, it often requires a long processing time. The results presented in \hyperref[fig.time]{Figure \ref*{fig.time}} and \hyperref[table.Salinas]{Table \ref*{table.Salinas}} verify the effectiveness of the proposed SMSB method in improving the performance of classification and reducing the processing time for Salinas image.

\subsection{Comparison with Deep Learning }
\begin{table}
  \centering
\caption{The quantitative classification results obtained for Indian Pines and Pavia University images for Proposed SMSB method and four deep learning based methods. The number of train and test data are chosen according to \cite{chen2016deep}. }
\resizebox{0.95\textwidth}{0.08\textheight}{%
\label{table.deep}
\begin{tabular}{c|c|c| c|c |c| c}
  \hline
  Dataset & Measurement & CNN & SSUN & SSRN & RPNet & SMSB \\ \hline
  \multirow{2}{*}{Indian Pines}& OA & 87.81 \!$\pm 0.32$ & 98.40 \!$\pm 0.37$& \textbf{98.80} \!$\pm 0.54$& 96.09 \!$\pm 0.66$& 97.89 \!$\pm 0.37$ \\
    & $\kappa$ & 85.30 \!$\pm 0.36$ & 98.14 \!$\pm 0.43$ & \textbf{98.61} \!$\pm 0.33$ & 95.46 \!$\pm 0.76$ & 98.01 \!$\pm 0.42$ \\ \hline
    \multirow{2}{*}{Pavia University}& OA & 92.28 \!$\pm 0.17$ & 99.46 \!$\pm 0.32$& \textbf{99.80} \!$\pm 0.51$& 99.34 \!$\pm 0.12$& 99.13 \!$\pm 0.26$ \\
    & $\kappa$ & 90.37 \!$\pm 1.01$ & 99.26 \!$\pm 0.44$ & \textbf{99.76} \!$\pm 0.71$ & 99.10 \!$\pm 0.17$ & 99.02 \!$\pm 0.74$ \\
  \hline
\end{tabular}
}
\end{table}

Recently, deep learning based methods have achieved remarkable performance in HSI classification. In this subsection, we conduct a brief comparison of our proposed SMSB method and a few of the state-of-the-art deep learning based methods. These methods include CNN \cite{chen2016deep}, SSUN \cite{xu2018spectral}, SSRN \cite{zhong2017spectral} and RPNet \cite{xu2018hyperspectral}. The CNN method exploits 3D CNNs with different regularization to extract spectral-spatial features. The SSUN method unifies LSTM and CNN models for spectral and spatial feature extraction, respectively. In SSRN method, a cascade of spectral and spatial residual networks are used to learn discriminative spectral-spatial features from 3D hyperspectral patches. The idea behind the RPNet method is to take random patches directly from the image, consider them as the convolutional kernels without training and hence reduce the processing time.

The experiments are conducted on Indian pines and Pavia University datasets. All the parameters of the aforementioned methods are chosen according to their corresponding papers. For all the deep learning based methods and the proposed SMSB method, the exact number of train and test data for Indian Pines and Pavia University image are chosen according to Table \RNum{1} and Table \RNum{2} of \cite{chen2016deep}, respectively.

\hyperref[table.deep]{Table \ref*{table.deep}} represents the quantitative results obtained for Indian Pines and Pavia University images. To keep this comparison concise only the overall accuracies (OA) and $\kappa$ coefficients are reported and the visual classification results are not considered. According to the obtained results, the SSRN method, which is based on residual networks, has outperformed the rest of the methods including the proposed SMSB method in both Indian Pines and Pavia University datasets. The SSUN method also outperforms the proposed method for both datasets but still the results are quite comparable.  Note that the main challenge of the deep learning based methods are their requirement for long processing times.

\section{Conclusion}
\label{section.five}
This paper presents a sparse modeling-based approach for hyperspectral image classification. Spectral blocks are used along with spatial groups to fully and jointly exploit spectral and spatial redundancies of the hyperspectral images. Using the spectral blocks, a sparse structure is designed and a dictionary is trained. The proposed method breaks the high-dimensional optimization problems of sparse modeling into several small-size sub-problems that are easier to tackle and faster to solve. The experimental results on three standard hyperspectral datasets and comparison with several state-of-the-art methods demonstrate the effectiveness of the proposed method for excellently classifying the hyperspectral data in short processing time. Future studies may be focused on developing different aspects of the proposed method. For instance, more advanced methods can be used to extract smarter spectral blocks or to improve the selection of the active spectral blocks.

\bibliographystyle{elsarticle-num}
\bibliography{Refs}

\end{document}